\documentclass[10pt,twocolumn,letterpaper]{article}

\usepackage[pagenumbers]{wacv}

\usepackage{multirow}
\usepackage{enumitem}
\usepackage{iftex}
\ifPDFTeX
  \usepackage[accsupp]{axessibility}
\fi

\definecolor{wacvblue}{rgb}{0.21,0.49,0.74}
\usepackage[breaklinks,colorlinks,allcolors=wacvblue]{hyperref}

\graphicspath{{./}{outputs/}}
\AtBeginDocument{\pdfcompresslevel=9 \pdfobjcompresslevel=2}

\title{Rad-R: A Raw-ADC Radar Dataset and Capture-Invariant SSM for Hardware-Fault Diagnosis}

\author{Mainak Mallick \quad Junghwan Yim \quad Alankrit Gupta \quad Seung-Kyum Choi\\
Georgia Institute of Technology\\
{\tt\small mmallick7@gatech.edu \quad jyim67@gatech.edu \quad agupta3129@gatech.edu \quad schoi@me.gatech.edu}
}

\begin{document}
\maketitle

\begin{abstract}
Automotive mmWave radar can develop vibration, antenna misalignment, radome blockage, and receive-channel degradation that corrupt the signal before perception begins. Data for these faults are scarce because each condition must be induced and measured on physical hardware. We introduce \textbf{Rad-R}, a raw-ADC dataset captured with a 4-chip 77\,GHz TI MMWCAS-RF-EVM cascade (192 virtual channels). Unlike existing raw-radar datasets, Rad-R pairs each recording with a controlled hardware fault at a calibrated severity, an independent physical severity measurement, and frame-synchronised IMU, temperature, GPS, and camera streams. Rad-R is a single-session dataset, so our generalisation claims are confined to a controlled cross-severity protocol in which train and test use physically distinct captures. A reproducible benchmark evaluates seven representative vision backbones and the proposed raw-IQ Mamba SSM (RadrNet) under within-clip, chirp-wise anytime, few-shot cross-capture, and controlled cross-severity protocols. Within-clip performance is near-saturated ($>0.98$ macro-F1), whereas cross-severity generalisation remains difficult: the absolute-phase RadrNet-DS falls to $0.49$ macro-F1. RadrNet-DS-CI replaces absolute phase with per-frame-standardised magnitude and relative chirp-to-chirp phase and ranks first on the controlled benchmark ($0.663$ vs. $0.628$ for the strongest RD-CNN; three seeds); the RadrNet family also leads on the anytime and few-shot budgets. A descriptive cross-modal analysis further finds that radar micro-Doppler covaries with independently measured IMU vibration energy (pooled Spearman $\rho=0.41$ across conditions). The complete dataset and code will be released publicly under permissive licences.

\end{abstract}

\section{Introduction}
\label{sec:intro}

Millimetre-wave (mmWave) frequency-modulated continuous-wave (FMCW) radar is central to automotive perception~\cite{radial2022,kradar2023,nuscenes2020} but degrades in deployment: vibration shifts alignment, partial radome blockage attenuates returns, individual Rx channels drift, and interference corrupts measurements. Existing public datasets either release raw analogue-to-digital-converter (ADC) samples under nominal operating conditions or provide processed tensors across varied scenes and weather~\cite{radial2022,kradar2023,cruw2021,coloradar2021,carrada2020}; none combines raw ADC with controlled hardware-fault annotations and a sensor-state evaluation suite. Rad-R fills that gap with (i) raw ADC from a 4-chip TI MMWCAS-RF-EVM cascade radar (Fig.~\ref{fig:deployment}; 12\,Tx $\times$ 16\,Rx = 192 virtual channels, 77\,GHz) under physically induced faults, (ii) per-radar-frame synchronised companion sensors (IMU, temperature $T$, GPS, camera), and (iii) a single-command evaluation harness with a Croissant-validated dataset card~\cite{croissant2024,datacards2022}. Because these faults act on the per-channel phase and gain that range--Doppler (RD) processing averages away, and because a diagnosis is needed early in each frame, we pair the dataset with RadrNet, a state-space model (SSM) that consumes the raw IQ tensor directly and can classify from a prefix of chirps. Hardware-fault data are also \emph{inherently scarce}: every example must be induced on real hardware. This constraint mirrors the limited-label regimes studied in fault diagnosis~\cite{zhang2019fewshot,wang2020fewshot} and industrial anomaly detection~\cite{bergmann2019mvtec}, motivating Rad-R as a benchmark for learning faults from limited labels.

\noindent\textbf{Contributions.}
\begin{enumerate}[leftmargin=*,itemsep=1pt,topsep=2pt]
\item \textbf{Rad-R dataset.} Nine captures ($\sim$27\,k frames; 324\,GB raw ADC) span healthy operation and four physically induced faults at two sensor-gated severities, with synchronised companion sensors, CC BY 4.0 licensing, and Croissant metadata.
\item \textbf{RadrNet.} A hierarchical 3-stage Mamba~\cite{mamba2023} processes raw complex in-phase/quadrature (IQ) samples with complex-aware encoding and multi-scale fusion; RadrNet-DS adds an RD-map CNN fused by an input-dependent gate.
\item \textbf{Capture-invariant RadrNet.} A capture-invariant encoding (per-frame-standardised magnitude plus relative chirp-to-chirp phase, which cancels constant global and per-channel phase offsets by construction) raises single-stream cross-severity F1 from $0.545$ to $0.612$; an ablation attributes most of this gain to magnitude standardisation (appendix). RadrNet-DS-CI exceeds the strongest RD-CNN by $+0.035$ mean macro-F1 and wins at each of three seeds.
\item \textbf{Controlled benchmark.} We compare RadrNet with seven vision backbones on two budget axes and a capture-disjoint cross-severity split. RadrNet-DS leads ResNet-18 at all six chirp budgets ($k=4$--$64$) and leads DeiT-III from five labelled target frames per class.
\end{enumerate}

\section{Related Work}
\label{sec:related}

\noindent\textbf{Radar datasets.} RADIal~\cite{radial2022} releases raw ADC with object boxes; K-Radar~\cite{kradar2023} provides 4D tensors across weather; CRUW~\cite{cruw2021}, RadarScenes~\cite{radarscenes2021}, RADIATE~\cite{oxfordrr2020}, and Dual Radar~\cite{dualradar2023} provide processed RA or point-cloud data; CARRADA~\cite{carrada2020} provides RAD tensors, and RODNet~\cite{rodnet2021} targets radar object detection. To our knowledge, no prior public radar dataset pairs raw ADC with hardware-fault labels (Table~\ref{tab:dataset_compare}).

\begin{table*}[tbp]
\centering
\caption{Comparison of public automotive / mmWave radar datasets: raw-ADC availability, MIMO virtual channels, data format, task, hardware-fault and severity annotations, and companion sensors.}
\label{tab:dataset_compare}
\footnotesize
\setlength{\tabcolsep}{4pt}
\resizebox{\textwidth}{!}{%
\begin{tabular}{lcccccccc}
\toprule
\textbf{Dataset} & \textbf{Year} & \textbf{Raw} & \textbf{MIMO} & \textbf{Format} & \textbf{Task} & \textbf{Fault} & \textbf{Sev.} & \textbf{Comp.\ sens.} \\
                 &      & \textbf{ADC} & \textbf{(virt.)} &           &       & \textbf{annot.} & & \\
\midrule
CARRADA~\cite{carrada2020}        & 2020 & --    & 12  & RAD tensor    & object detect. & --   & --  & cam \\
CRUW~\cite{cruw2021}              & 2021 & --    & 12  & RA map        & object track.  & --   & --  & cam \\
RadarScenes~\cite{radarscenes2021}& 2021 & --    & n/r & point cloud   & semantic seg.  & --   & --  & cam, GPS \\
RADIal~\cite{radial2022}          & 2022 & \checkmark & 192 & ADC + RD + RA & object detect. & --   & --  & cam, lidar, GPS \\
K-Radar~\cite{kradar2023}         & 2023 & --    & n/r & 4D tensor     & object detect. & --$^{\ddag}$ & -- & cam, lidar, GPS \\
Dual Radar~\cite{dualradar2023}   & 2024 & --    & 2304 & 4D tensor     & object detect. & --   & --  & cam, lidar \\
\midrule
\textbf{Rad-R (ours)}             & 2026 & \checkmark & 192 & ADC + RD + RA & \textbf{fault classification} & \textbf{4 types} & \textbf{2} & cam, IMU, GPS, $T$, $RH$ \\
\bottomrule
\end{tabular}}
\\[2pt]
{\footnotesize Virtual channels = nominal Tx$\times$Rx of the sensor MIMO array (\emph{n/r}: not an imaging-MIMO array, or not reported); only RADIal and Rad-R release the raw per-channel ADC needed to access it. RA / RD / RAD: range--azimuth / range--Doppler / range--azimuth--Doppler; cam: camera; $T$ / $RH$: temperature / relative humidity. $^{\ddag}$\,K-Radar varies environmental conditions (rain, snow, fog) but does \emph{not} induce hardware faults at the sensor.}
\end{table*}

\noindent\textbf{Radar fault, interference, and raw-ADC learning.} Degraded sensing is safety-relevant in deployed ADAS~\cite{nhtsa_radar2023}, yet no public benchmark exposes how controlled radar hardware faults propagate into learned perception. This gap is analogous to the distribution shifts studied by WILDS~\cite{koh2021wilds}, but at the sensor-hardware level. RIMformer~\cite{rimformer2024} and RIME-Net~\cite{rimenet2023} target interference mitigation on RD maps; FMCW degradation studies~\cite{fdiir2023,linge2020vibration,roos2024alignment} catalogue faults without a public raw-ADC benchmark; bearing-style diagnosis~\cite{cwru_bearing} does not model radar's complex multi-channel structure. ADCNet~\cite{adcnet2023} distils RD$\to$ADC; Mamba-RODNet~\cite{mambarod2024} applies Mamba to RD; and T-FFTRadNet~\cite{tfftradnet2024} uses transformers on radar tensors. To our knowledge, RadrNet is the first hierarchical Mamba evaluated on complex raw IQ for both chirp-wise anytime and few-shot label-budget inference.

\noindent\textbf{Streaming and anytime radar inference.} SSMRadNet~\cite{ssmradnet2026} processes raw ADC samples with sample-wise and chirp-wise SSMs~\cite{ssmgu2022}, while RadMamba~\cite{radmamba2025} applies a Mamba SSM to radar micro-Doppler activity recognition. RECORD~\cite{record2024} uses a recurrent CNN for causal online detection, and RAVEN~\cite{raven2025} formalises \emph{chirp-wise early exit} with per-budget calibration. These works motivate our anytime / chirp-wise evaluation (Section~\ref{subsec:anytime}). We adopt the paired-bootstrap and McNemar protocol of~\cite{bouthillier2025paired} for rigorous multi-seed comparisons.

\section{The Rad-R Dataset}
\label{sec:data}
\label{sec:dataset}

\subsection{Hardware and Capture Setup}
\label{sec:hw}

\begin{figure*}[t]
  \centering
  \includegraphics[width=\textwidth]{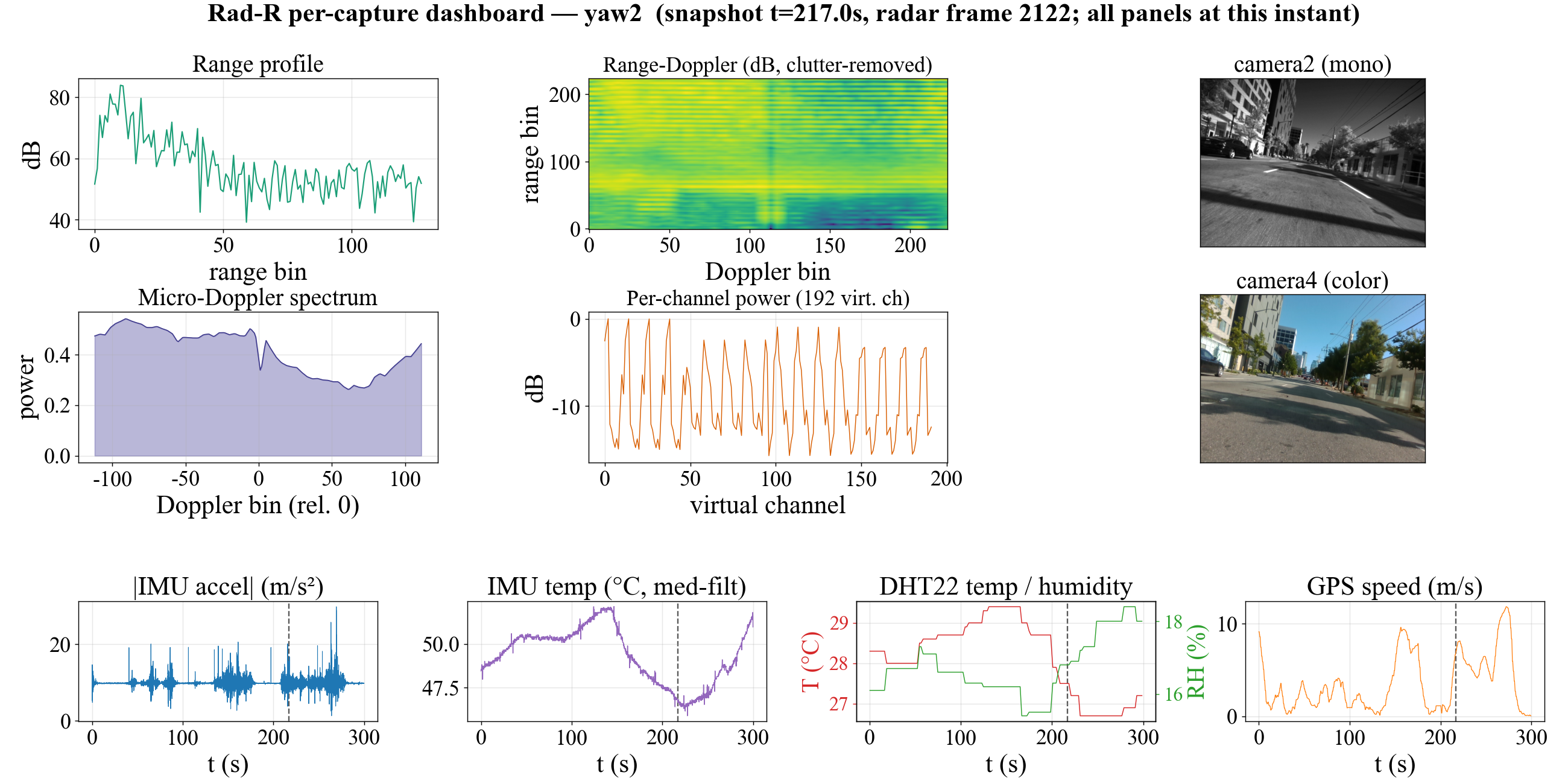}
  \caption{\textbf{Per-capture dashboard} for the severe-misalignment capture (\texttt{yaw2}, $10^\circ$ yaw) at radar frame 2122 ($t=217$\,s). \textbf{Top left:} chirp-mean range profile, range--Doppler map (static-clutter cancellation for display only), micro-Doppler spectrum, and power across the 192 virtual channels at this frame. \textbf{Top right:} the two synchronised camera streams at the same timestamp (faces and licence plates blurred). \textbf{Bottom:} IMU acceleration magnitude, IMU temperature, DHT22 temperature/humidity, and GPS speed over the 5-minute capture; the dashed line marks this frame. Dashboards for the other eight captures are in Appendix~\ref{app:dashboards}.}
  \label{fig:dashboard}
\end{figure*}

Rad-R uses a TI MMWCAS-RF-EVM~\cite{ti_mmwcas} (Fig.~\ref{fig:deployment}; an exploded view of the rig is in Appendix~\ref{app:induction}): a 4-chip AWR1243 cascade at 77\,GHz with 12\,Tx and 16\,Rx, yielding 192 virtual channels. Four DCA1000EVMs record the radar stream~\cite{ti_dca1000}. A $1/16$-inch polypropylene radome is installed for \emph{all} captures. A BNO055 IMU ($\sim$33\,Hz), two DHT22 temperature probes, u-blox NEO-M9N GPS, and two RealSense D435 camera streams (640$\times$480 at $\sim$30\,fps) are aligned to the radar frame index in synchronised HDF5 files. Figure~\ref{fig:dashboard} shows one radar frame of a misalignment capture together with every companion stream at that instant.

\noindent\textbf{Chirp configuration and virtual-array geometry.} Every Rad-R capture uses TI's reference cascade-imaging chirp parameters (full table in the appendix): 256 ADC samples per chirp at an 8\,Msps (mega-samples per second) sample rate, a 78.986\,MHz/$\mu$s ramp giving 2.49\,GHz effective bandwidth (0.060\,m range resolution, 15.2\,m maximum range), 64 loops per frame and a 100\,ms frame period ($\sim$10\,Hz). The cascade runs in \emph{TDM-MIMO} mode: within each loop the 12 transmit antennas are time-multiplexed across the 16 physical receive channels, so MIMO unmixing reconstructs a 192-element virtual array ($16\times12$) whose angular resolution far exceeds that of any single chip. This dense virtual array carries independent per-receiver phase and amplitude structure in raw IQ that incoherent range--Doppler averaging discards, which is what makes the raw-IQ representation richer than the averaged RD map for fault classification. The four DCA1000EVMs stream raw \texttt{int16} ADC samples (real/imag interleaved) to disk in TI's standard \texttt{*\_data.bin}/\texttt{*\_idx.bin} format (one master, three slaves per session).

\begin{figure}[t]
  \centering
  \includegraphics[width=\linewidth]{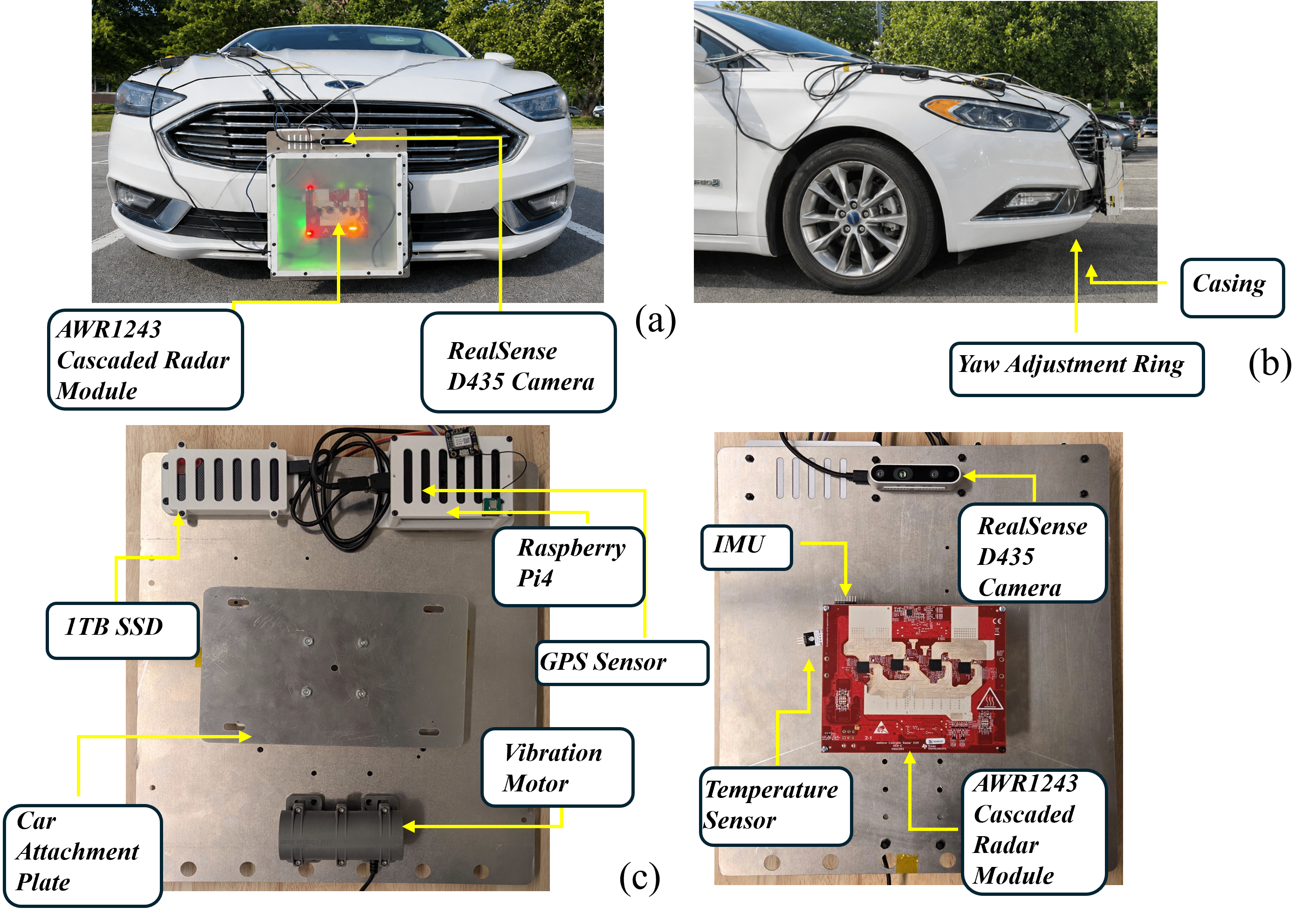}
  \caption{Rad-R in deployment. \textbf{Top:} the rig mounted on the test vehicle (front and side), behind the front grille. \textbf{Bottom:} the assembled sensor plate (front and back) showing the AWR1243 cascade radar, RealSense camera, BNO055 IMU, GPS, eccentric-mass vibration motor, temperature probe, Raspberry Pi and 1\,TB SSD. Captures are collected on private lots and permitted campus roads at low speed.}
  \label{fig:deployment}
\end{figure}

\subsection{Fault Taxonomy and Induction Procedures}
\label{sec:taxonomy}

\begin{figure*}[t]
  \centering
  \includegraphics[width=\textwidth]{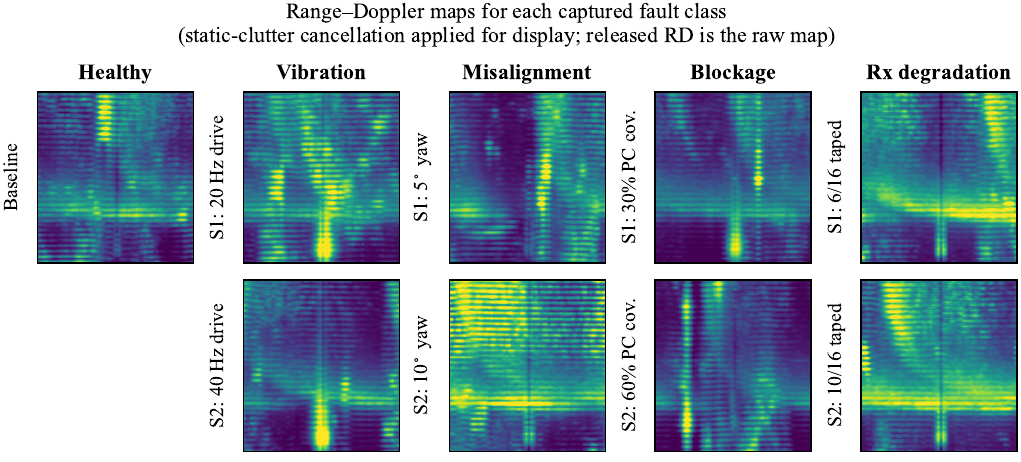}
  \caption{Range--Doppler maps for each fault $\times$ severity cell (columns: fault type; rows: severity). For display we apply static-clutter cancellation (slow-time mean subtraction) so scene structure is visible rather than the zero-Doppler leakage line; the released RD product is the raw map. The per-fault signatures are physically interpretable but subtle (Sec.~\ref{sec:stats}).}
  \label{fig:rd_grid}
\end{figure*}

\begin{table}[tbp]
\centering
\caption{Fault taxonomy and induction parameters captured in Rad-R.}
\label{tab:taxonomy}
\footnotesize
\setlength{\tabcolsep}{3pt}
\begin{tabular}{lllll}
\toprule
\textbf{Fault} & \textbf{Induction} & \textbf{S1} & \textbf{S2} & \textbf{Severity gate} \\
\midrule
Healthy           & n/a                  & n/a            & n/a            & n/a  \\
Vibration         & Ecc.\ DC motor       & 20\,Hz         & 40\,Hz         & IMU RMS \\
Misalign.         & Yaw jig              & 5$^\circ$ & 10$^\circ$ & Inclinometer \\
Blockage          & PC on radome         & 30\%  & 60\%  & Reflector dB \\
Rx degrad.        & Cu foil tape         & 6/16 Rx        & 10/16 Rx       & Cal sweep \\
\bottomrule
\end{tabular}
\end{table}

Four physically induced fault types are each captured at two severity levels (S1 mild, S2 severe), plus a healthy baseline (S0); fault induction is hardware-physical, not synthetic injection (Table~\ref{tab:taxonomy}). Each fault's severity is gated by an \emph{independent physical sensor}, not by the operator's nominal setting: an eccentric-mass DC motor bolted to the radar bracket drives \emph{vibration} at 20/40\,Hz with the BNO055 IMU recording ground-truth tri-axial acceleration; a precision yaw jig with $1^\circ$ graduations sets \emph{misalignment} to $5^\circ$/$10^\circ$, verified with a $0.1^\circ$ digital inclinometer; a laser-cut clear polycarbonate sheet over the always-installed polypropylene radome produces 30\%/60\% \emph{blockage}, calibrated against a corner-reflector dB drop; and 3M~1181 copper foil tape covering 6/16 or 10/16 Rx antenna patches induces \emph{Rx-channel degradation}, with the per-channel dB drop confirmed by a single-tone calibration sweep before each capture. Because the induction is physical and the gate independent, the severity labels are objective rather than nominal. Thermal stress and RF interference are part of the dataset specification but deferred to future releases.

\subsection{Capture Protocol and Dataset Statistics}
\label{sec:stats}

Each capture is a continuous 5-minute recording at $\sim$10\,fps ($\sim$3\,k frames); the fault state is held throughout. Rad-R contains 9 capture runs ($\sim$27\,k frames, $\sim$324\,GB raw ADC). For ML training we release a pre-processed cache of 1\,800 frames (200 evenly sampled per capture, class-balanced 200 healthy / 400 per fault) with 224$\times$224 RD maps (dB-scaled, per-frame normalised) and raw IQ at the canonical (loops, samples, virtual channels) layout. Companion sensors carry physically meaningful per-fault signatures (e.g.\ vibration $1.5\times$ higher tri-axial gyro RMS; per-capture traces in Appendix~\ref{app:visualizations}), enabling severity regression and sensor-fusion research.

\noindent\textbf{Signal-processing pipeline.} The released RD maps are produced from raw ADC by a standard pipeline: windowed range/Doppler FFTs, non-coherent integration across the 192 virtual channels, dB conversion, per-frame normalisation, and a bilinear resize to $224\times224$~\cite{harris1978,richards2014} (recipe in the released code and dataset card). Range--azimuth and micro-Doppler maps are derived analogously and shipped for completeness, although the benchmark uses only RD.

\noindent\textbf{Fault signatures in the released representations.} Figure~\ref{fig:rd_grid} shows representative range--Doppler maps for every fault $\times$ severity cell. The visual signatures are physically interpretable but subtle: vibration produces broadband Doppler spreading; blockage attenuates returns and elevates the noise floor; Rx-channel degradation perturbs the sidelobe structure while leaving a recognisable peak. Misalignment is near-invisible in RD: a few-degree yaw leaves range and Doppler essentially unchanged, acting only on the azimuth and per-channel phase that the channel average discards. This subtlety is precisely what motivates learned classifiers over hand-crafted features, and what makes the raw-IQ representation (which retains the per-channel structure averaged out of the RD map) valuable.

\subsection{Evaluation Splits}
\label{sec:splits}

Rad-R supports four protocols:
\begin{enumerate}[leftmargin=*,label=(\arabic*),itemsep=2pt,topsep=2pt]
\item \textbf{Within-clip 80/20 temporal block.} The first 80\% of each capture trains the model and the last 20\% tests it. The split contains no shared radar frames, but it does share scene, sensor, and configuration. We therefore treat it as a confounded upper bound rather than an independent generalisation test~\cite{radial2022,coloradar2021}.
\item \textbf{Anytime chirp budget.} Models train on full 64-chirp frames from protocol (1) and are evaluated using only the first $k\in\{4,8,16,32,48,64\}$ chirps (Section~\ref{subsec:anytime}). This tests architectural behavior under truncated evidence, but inherits the within-capture confound.
\item \textbf{Few-shot cross-capture adaptation.} A source-trained model receives $k\geq1$ labelled frames per fault class from a held-out target capture and is evaluated on the remaining target frames (Section~\ref{subsec:fewshot}).
\item \textbf{Controlled cross-severity 2-fold.} Training uses one severity of each fault and testing uses the other (Section~\ref{subsec:crosssev}); train and test therefore use physically distinct captures for every fault.
\end{enumerate}
The single-day, single-device scope does not support rigorous held-out day, device, or scene evaluation; these axes are reserved for future releases.

\section{Models and RadrNet}
\label{sec:models}

\begin{figure*}[t]
\centering
\includegraphics[width=\textwidth]{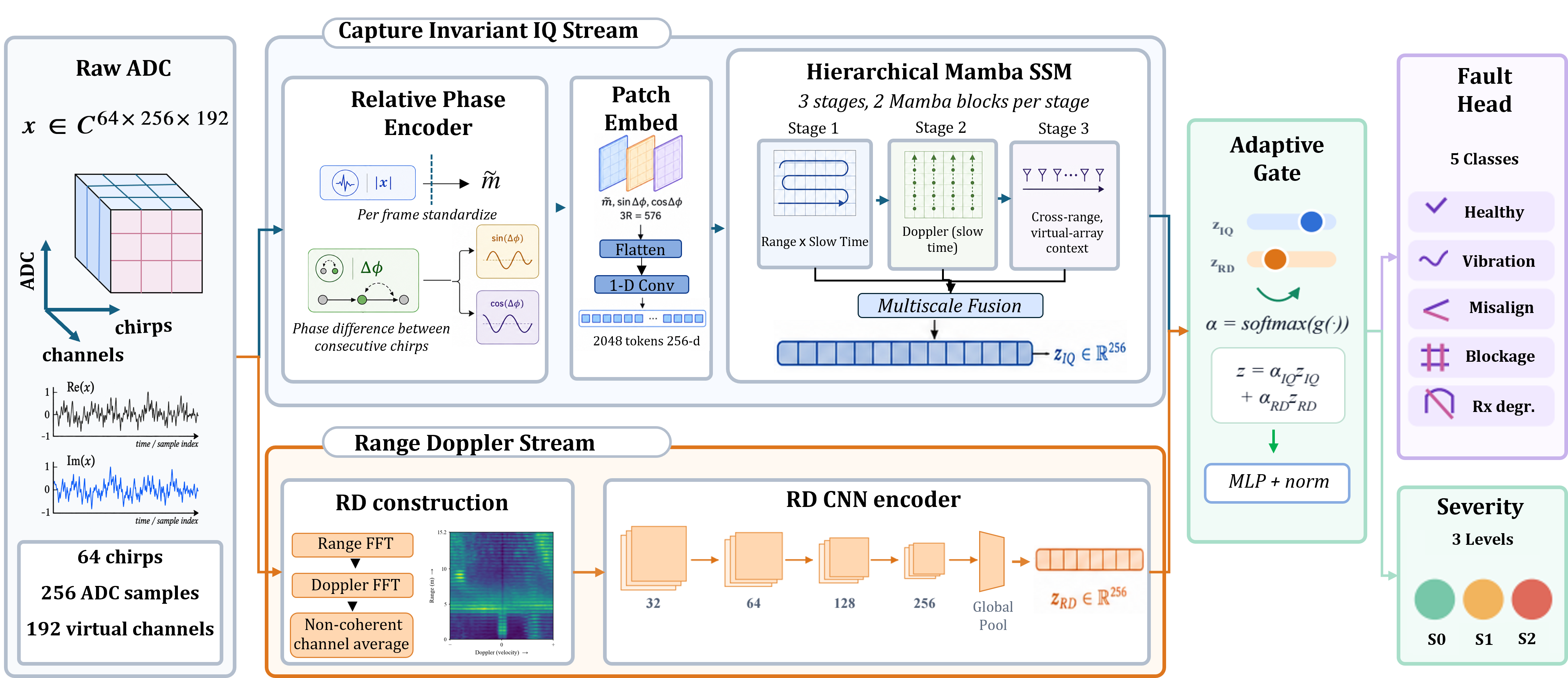}
\caption{\textbf{RadrNet-DS-CI architecture.} The capture-invariant IQ stream encodes per-frame-standardised magnitude $\tilde m$ and relative chirp-to-chirp phase $(\sin\Delta\phi,\cos\Delta\phi)$, then processes range, Doppler, and virtual-array structure with a 3-stage hierarchical Mamba~\cite{mamba2023} and multi-scale fusion. An input-dependent 2-way gate combines the IQ representation with a compact RD-CNN representation before the 5-way fault and 3-way severity heads. RadrNet-CI omits the RD stream.}
  \label{fig:radrnet_arch}
\end{figure*}

We benchmark models spanning the representation axis (RD map vs.\ raw complex IQ) and the inductive-bias axis (CNN, transformer~\cite{vaswani2017attention}, SSM): seven published RD-map backbones (\textbf{ResNet-18}, 11.2\,M~\cite{he2016deep}; \textbf{ViT-Small}, 21.5\,M~\cite{dosovitskiy2020image}; \textbf{ConvNeXt-V2}~\cite{convnextv2}; \textbf{DeiT-III}~\cite{deit3}; \textbf{Swin-T}~\cite{swin}; \textbf{PVTv2}~\cite{pvtv2}; \textbf{EfficientNetV2}~\cite{efficientnetv2}), and the proposed \textbf{RadrNet} family. Every RD backbone uses a 1-channel stem and dual fault/severity heads. Severity is a gate level (S0/S1/S2) common to all faults, so a single shared severity head is used; it is trained as an auxiliary loss (weight 0.5; appendix) and does not enter the fault metrics.

\noindent\textbf{RadrNet} is a hierarchical 3-stage Mamba~\cite{mamba2023,mamba2_2024} on raw complex IQ tensors of shape $(B,2,C,S,R)$: batch, real/imaginary part, chirps $C$, ADC samples per chirp $S$, and virtual channels $R$. It patch-embeds a $4R$-channel encoding (real, imaginary, magnitude, absolute phase) along $C\!\cdot\!S$, then scans joint range/slow-time, Doppler, and cross-channel virtual-array structure. Multi-scale stage features feed dual fault and severity heads (Fig.~\ref{fig:radrnet_arch}). Four train-only augmentations (global phase rotation, per-Rx amplitude jitter, Rx dropout, and per-frame standardisation) randomise capture-specific phase and gain. \textbf{RadrNet-DS} ($\sim$5.1\,M) adds a compact RD-map CNN and fuses the IQ and RD representations with a 2-way input-dependent gate.

\noindent\textbf{Capture-invariant SSM.} On the controlled cross-severity split (Section~\ref{subsec:crosssev}) the raw-IQ models above are the weakest cross-severity generalisers, despite leading within-clip. Absolute phase exposes global and per-channel calibration offsets that remain nearly constant within a capture; per-frame magnitude scale can provide an additional capture cue. \textbf{RadrNet-CI} ($\sim$4.1\,M) keeps the 3-stage Mamba backbone but replaces real/imag$+$absolute-phase encoding with (i) per-frame-standardised magnitude and (ii) relative chirp-to-chirp phase $(\Delta\phi)$, encoded as $(\sin\Delta\phi,\cos\Delta\phi)$ ($3R$ input channels). Differencing cancels constant global and per-channel phase offsets, while standardisation removes global magnitude scale; Section~\ref{subsec:crosssev} and the appendix isolate the two components. \textbf{RadrNet-DS-CI} ($\sim$4.7\,M) fuses this invariant-IQ branch with a compact RD-CNN through the same 2-way input-dependent gate.

\noindent\textbf{Formulation.} Let $x\in\mathbb{C}^{C\times S\times R}$ be one frame ($C{=}64$ chirps, $S{=}256$ ADC samples, $R{=}192$ virtual channels).

\noindent\emph{Encoding.} The capture-invariant encoding is the parameter-free map
\begin{align}
E(x)&=\big[\tilde m,\ \sin\Delta\phi,\ \cos\Delta\phi\big]\in\mathbb{R}^{3\times C\times S\times R},\label{eq:enc}\\
\tilde m&=\frac{|x|-\mu_x}{\sigma_x},\quad
\Delta\phi_{c,s,r}=\mathrm{wrap}\big(\angle x_{c+1,s,r}-\angle x_{c,s,r}\big),\nonumber
\end{align}
with $\mu_x,\sigma_x$ taken over all $(c,s,r)$, $\mathrm{wrap}(\cdot)$ to $(-\pi,\pi]$ and zero padding at $c{=}0$. For $x'_{c,s,r}=g\,e^{j(\phi_0+\psi_r)}x_{c,s,r}$ (global gain $g>0$, global phase $\phi_0$, per-channel calibration phase $\psi_r$), $E(x')=E(x)$ exactly; a per-channel gain $g_r$ is not cancelled and is covered by the training-time Rx-amplitude jitter.

\noindent\emph{Tokens and SSM stages.} A strided 1-D convolution over the flattened chirp$\times$sample axis yields
\begin{equation}
u_n=W_e\,\mathrm{vec}\big(E(x)_{[\,:\,,\,8n:8n+8]}\big)+b_e\in\mathbb{R}^{d},\quad n=1,\dots,N,
\label{eq:tok}
\end{equation}
with $N=CS/8=2048$ and $d=256$, so all $3R$ encodings of eight neighbouring samples enter each token. Each stage stacks two pre-norm residual selective-SSM blocks~\cite{mamba2023},
\begin{equation}
h\leftarrow h+\mathrm{SSM}(\mathrm{LN}(h)),\qquad s_t=\bar A_t s_{t-1}+\bar B_t h_t,\ \ y_t=C_t s_t,
\label{eq:ssm}
\end{equation}
with input-dependent $(\Delta_t,B_t,C_t)$. Stage~1 scans $u_{1:N}$ (joint range$\times$slow time); its output is reshaped to a $C\times S/8$ grid and stage~2 scans along $c$ for each range patch (Doppler); stage~3 scans the $W_p$-projected per-patch means across range patches (cross-range, virtual-array context). With $f_k$ the token mean after stage $k$,
\begin{align}
\beta&=\mathrm{softmax}\big(\mathrm{MLP}([\mathrm{LN}f_1;\mathrm{LN}f_2;\mathrm{LN}f_3])\big)\in\mathbb{R}^{3},\label{eq:fuse}\\
z_{\mathrm{IQ}}&=\mathrm{MLP}\Big(\textstyle\sum_{k}\beta_k\,\mathrm{LN}f_k\Big).\nonumber
\end{align}

\noindent\emph{RD stream and gate.} The RD stream uses the released map
\begin{equation}
D(x)=\mathrm{resize}_{224}\Big(10\log_{10}\tfrac{1}{R}\textstyle\sum_{r}\big|\mathcal{F}_{c}\mathcal{F}_{s}(w\odot x)\big|^{2}_{r}\Big)
\label{eq:rd}
\end{equation}
(Hann windows $w$, min--max normalised per frame) and $z_{\mathrm{RD}}=\mathrm{CNN}(D(x))\in\mathbb{R}^{d}$ (four strided convolutions, $32\!\to\!64\!\to\!128\!\to\!256$ channels, global average pool). An input-dependent gate $\alpha=\mathrm{softmax}\big(\mathrm{MLP}([\mathrm{LN}z_{\mathrm{RD}};\mathrm{LN}z_{\mathrm{IQ}}])\big)\in\mathbb{R}^{2}$ fuses the two streams,
\begin{equation}
z=\mathrm{LN}\big(\mathrm{GELU}(W_o(\alpha_{\mathrm{RD}}\,\mathrm{LN}z_{\mathrm{RD}}+\alpha_{\mathrm{IQ}}\,\mathrm{LN}z_{\mathrm{IQ}}))\big),
\label{eq:gate}
\end{equation}
and two linear heads on $z$ give the 5-way fault and 3-way severity logits, trained with $\mathcal{L}=\mathrm{CE}_{\mathrm{fault}}+0.5\,\mathrm{CE}_{\mathrm{sev}}$. RadrNet-CI feeds $z_{\mathrm{IQ}}$ to the heads directly.

\section{Experiments and Results}
\label{sec:experiments}

\noindent\textbf{Setup.} All models train on the same 1{,}800-frame cache with AdamW (lr~$10^{-4}$, weight decay $10^{-4}$), cosine schedule, gradient-norm clip 1.0, batch size 16, and 20 epochs (a subset of the cross-severity runs use 25; see appendix). The saved runs select the epoch with maximum macro-F1 on the evaluation split; because no separate validation partition was used, the within-clip and cross-severity scores are oracle-selected upper bounds. Few-shot adaptation instead uses a fixed 30-step schedule. Fault macro-F1 is the primary metric; fault/severity accuracy and 15-bin ECE are retained in the released result records. Full hyperparameters, metrics, and compute are in the appendix.

\subsection{Within-Clip Split as Context}
\label{sec:withinclip}

\noindent\textbf{Within-clip split is near-saturated.} The within-clip 80/20 split is the regime on which the anytime models are trained (Section~\ref{sec:splits}), but it does not discriminate architectures: under the identical recipe of Section~\ref{sec:models}, the strongest published backbones reach $0.974$ macro-F1 (DeiT-III, Swin-T) and minimal-inductive-bias raw-IQ baselines (a $0.7$\,M 1D-CNN, a $6.8$\,M IQ-transformer) reach $0.989$--$0.994$. Because this split shares scene and sensor, we report it only as a contextual upper bound. Generalisation claims rely on the cross-capture and controlled cross-severity protocols; the anytime experiment tests behavior under a chirp budget within the same confounded split. The released records additionally include 15-bin ECE~\cite{guo2017calibration}, which matters when confidence influences a safety decision.

\subsection{Anytime / Chirp-Wise Inference}
\label{subsec:anytime}

\noindent\textbf{Protocol.} The chirp-budget axis compares RadrNet-DS with a ResNet-18 RD-map baseline. Both train on full 64-chirp frames from the within-clip 80/20 split (Section~\ref{sec:splits}) for 20 epochs at each of five seeds. At evaluation, each model receives only the first $k\in\{4,8,16,32,48,64\}$ chirps: RadrNet-DS consumes the truncated raw IQ directly, whereas ResNet-18 receives a fixed-size RD input recomputed after chirps $k,\ldots,63$ are set to zero. Significance follows \cite{bouthillier2025paired}: a one-sided paired bootstrap ($n{=}2000$, $H_0$: $\mathrm{F1}_{\mathrm{RadrNet\text{-}DS}} \leq \mathrm{F1}_{\mathrm{ResNet}}$) and a two-sided McNemar exact test on 1\,800 pooled predictions (five seeds $\times$ 360 frames) at each $k$. Figure~\ref{fig:two_axis} (left) plots the curves; the full table is in the appendix.

\noindent\textbf{Motivation.} The protocol builds on recent efficient radar sequence models: sample-wise and chirp-wise SSM processing~\cite{ssmradnet2026}, micro-Doppler-oriented Mamba modeling~\cite{radmamba2025}, causal online recurrent detection~\cite{record2024}, and calibrated chirp-wise early exit~\cite{raven2025}.

\begin{figure*}[t]
  \centering
  \includegraphics[width=\textwidth]{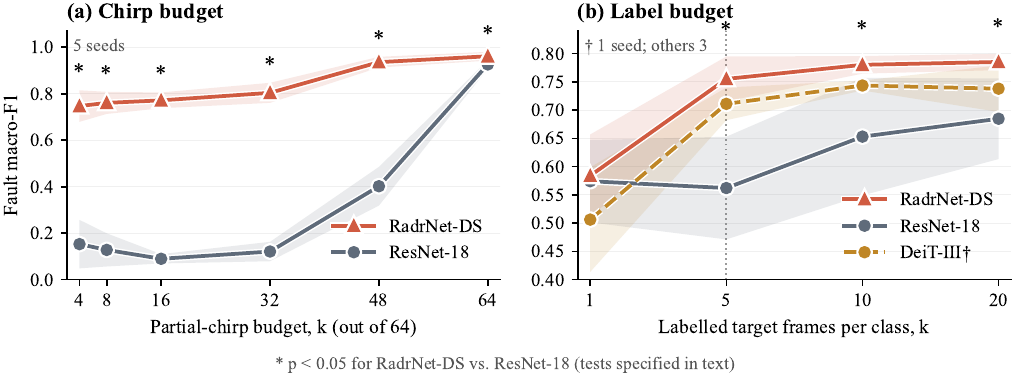}
  \caption{Two budget axes (mean $\pm$ std bands). \textbf{(Left) Chirp budget (anytime).} Within-clip 80/20 MIMO split, five seeds; stars mark paired-bootstrap $p<0.05$ at each $k$ (test defined in the text). \textbf{(Right) Label budget (few-shot).} Cross-capture adaptation; stars mark unpaired-bootstrap $p<0.05$ for RadrNet-DS over the ResNet-18 RD-CNN.}
  \label{fig:two_axis}
\end{figure*}

\noindent\textbf{Findings.} RadrNet-DS attains higher F1 than ResNet-18 at every $k$ (paired-bootstrap $p<0.05$ throughout). The gap is largest at small $k$ ($\Delta$F1 stays in $[+0.594,+0.683]$ for $k\leq32$, where ResNet-18 sits near the majority-class macro-F1) and narrows to $+0.035$ at the full $k{=}64$ frame, still significant (McNemar $p=5.9{\times}10^{-5}$). This pattern is consistent with the selective scan integrating truncated slow-time evidence more effectively than an RD-CNN operating on a poorly resolved, zero-padded Doppler spectrum; it does not by itself isolate the causal contribution of any single component~\cite{ssmradnet2026,radmamba2025,record2024,raven2025}.

\noindent\textbf{Latency.} On an RTX A4000 (fp32), RadrNet-DS takes $5.5$--$6.2$\,ms/frame and ResNet-18 takes $2.1$--$4.6$\,ms/frame. The roughly $2\times$ wall-time buys $+0.5$--$0.7$ macro-F1 at small $k$ while remaining below the 100\,ms frame deadline.

\subsection{Few-Shot Cross-Capture Adaptation}
\label{subsec:fewshot}

\noindent\textbf{Protocol.} A complementary \emph{label-budget} axis to the chirp budget of Section~\ref{subsec:anytime}. Each base model is trained on the source (one-severity) captures for 15 epochs; at test we draw $k$ labelled frames \emph{per fault class} from the held-out target capture, fine-tune for 30 steps at $\mathrm{lr}{=}10^{-4}$ (identical recipe across models), and evaluate on the remaining target frames. We sweep $k\in\{1,5,10,20\}$ over 2 folds $\times$ 3 support draws per seed, comparing RadrNet-DS against ResNet-18 (the canonical RD-CNN) and DeiT-III (the strongest published backbone). As models draw \emph{independent} support sets, significance uses a one-sided \emph{unpaired} bootstrap ($n{=}2000$). RadrNet-DS and ResNet-18 use three seeds (18 fold$\times$draw observations per $k$); DeiT-III and the capture-invariant variants are single-seed. The full per-budget table is in the appendix.

\noindent\textbf{Findings.} At $k{=}1$ RadrNet-DS is statistically tied with the ResNet-18 RD-CNN ($0.584$ vs. $0.574$, $p{=}0.335$). From $k{=}5$ it leads both competitors ($0.755$ vs. ResNet's $0.562$, $p<10^{-3}$, and DeiT-III's single-seed $0.711$). Its margin over DeiT-III is modest ($+0.04$--$0.05$ F1). From $k{=}5$ to 20, RadrNet-DS-CI rises from $0.732$ to $0.761$, while three-seed RadrNet-DS rises from $0.755$ to $0.785$; the invariant variant trades some few-shot adaptability for cross-severity gains.

\subsection{Controlled Cross-Severity Generalisation}
\label{subsec:crosssev}
\begin{figure}[tbp]
  \centering
  \includegraphics[width=\linewidth]{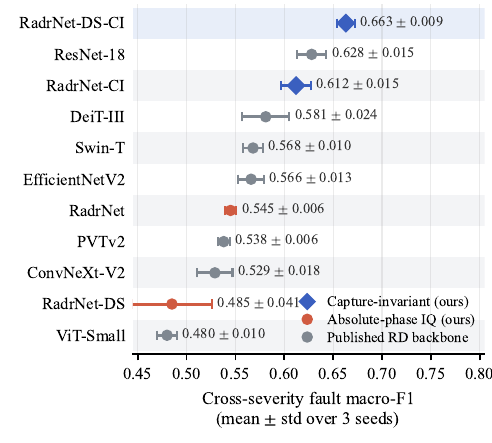}
  \caption{Controlled cross-severity 2-fold benchmark: fault macro-F1 per model (three-seed mean; thin caps are std over seeds). Colour distinguishes our RadrNet variants (capture-invariant; absolute-phase raw-IQ) from off-the-shelf backbones. Per-model values with input modality are tabulated in the appendix.}
  \label{fig:crosssev_rank}
\end{figure}

\noindent\textbf{Protocol.} The within-clip and anytime protocols share scene and sensor between train and test. To control this shortcut~\cite{geirhos2020shortcut}, we train on one severity of each fault and test on the other (2 folds), so every fault uses physically distinct train and test captures~\cite{koh2021wilds}. \textbf{Result.} Figure~\ref{fig:crosssev_rank} reports fault macro-F1: off-the-shelf baselines fall from $0.71$--$0.97$ within-clip to $\leq0.63$, and the ranking inverts. RD-map CNNs lead the absolute-phase raw-IQ models (RadrNet $0.545$, RadrNet-DS $0.485$). The capture-invariant encoding raises the single-stream RadrNet-CI to $0.612\pm0.015$, while RadrNet-DS-CI ranks first at $0.663\pm0.009$, $+0.035$ above ResNet-18 ($0.628\pm0.015$) and higher at each of three seeds. The largest gain occurs in the severe$\to$mild fold (appendix). An encoding ablation (Appendix~\ref{app:ci_ablation}) shows that magnitude standardisation alone reaches $0.631\pm0.028$ and relative phase alone $0.399\pm0.025$: relative phase does not raise the mean but lifts the harder severe$\to$mild fold ($0.538\to0.578$) through blockage and Rx-degradation recall, and the per-fault breakdown there also explains why the RD branch hurts the absolute-phase model ($0.545\to0.485$) yet helps the invariant one ($0.612\to0.663$). Substantial headroom remains: more than $0.3$ F1 is lost relative to within-clip performance. A descriptive cross-modal analysis in the appendix shows that radar micro-Doppler covaries with independently measured IMU vibration energy across the three captured conditions (pooled Spearman $\rho=0.41$).

\noindent\textbf{Anytime under cross-severity.} Under partial-chirp budgets (full values in the appendix), the two invariant models trade off: RadrNet-DS-CI is substantially weaker at small budgets ($0.30$ at $k{=}4$, $0.601$ at $k{=}64$), whereas RadrNet-CI reaches $0.479$ from only $4$ of $64$ chirps. The $0.601$ endpoint comes from the separate prefix-evaluation harness with flat seed$\times$fold averaging and final-epoch weights, and is therefore not interchangeable with the $0.663$ full-frame aggregate above; it is, however, a selection-free estimate, and at $k{=}64$ the ordering is unchanged: RadrNet-DS-CI $0.601$, ResNet-18 $0.591$, RadrNet-CI $0.580$, RadrNet-DS $0.441$ (three seeds for the CI models, one for the other two; Table~\ref{tab:ci_anytime}). Neither invariant model dominates across budgets; the absolute-phase RadrNet-DS loses its within-clip anytime advantage and trails ResNet-18 at $k{=}64$.

\section{Conclusion}
\label{sec:conclusion}

Rad-R is the first radar dataset to pair raw ADC with controlled, physically induced hardware faults at independently gated, calibrated severities and per-frame synchronised companion sensors. Its controlled cross-severity benchmark exposes a failure that within-clip evaluation hides: absolute-phase raw-IQ SSMs are vulnerable to capture-specific shortcuts and decline sharply across severities. The capture-invariant encoding removes constant phase offsets and global gain by construction---with magnitude standardisation, not relative phase, accounting for most of the gain---and RadrNet-DS-CI is the strongest model on the controlled challenge ($0.663$ vs. $0.628$ for the best RD-CNN). Substantial headroom remains ($>0.3$ F1 below the within-clip regime), and the capture-invariant models trade some calibration for this robustness (ECE $\approx0.17$--$0.21$ vs. $\approx0.14$ for ResNet-18), so they require recalibration before deployment. Rad-R is single-session, and the saved within-clip and cross-severity checkpoints are oracle-selected on their evaluation splits; both limitations make the current numbers upper bounds rather than deployment estimates. The complete dataset and code will be released publicly under permissive licences.


\section*{Acknowledgements}
This work was supported by the Korea Evaluation Institute of Industrial Technology (KEIT) grant funded by the Ministry of Trade, Industry \& Energy (MOTIE) of the Republic of Korea (No.~RS-2024-00448797).

\clearpage
\appendix
\setcounter{figure}{0}\setcounter{table}{0}
\renewcommand{\thefigure}{S\arabic{figure}}
\renewcommand{\thetable}{S\arabic{table}}
\twocolumn[%
  {\centering\LARGE\bfseries Supplementary Material\par}
  \vspace{1em}
  {\normalsize\noindent The following appendices contain the supplementary material referenced in the main text; figures and tables carry an ``S'' prefix, while references are shared with the main text.\par}
  \vspace{1.5em}
]

\section{Detailed Radar Configuration}
\label{app:hwconfig}
\label{app:config}

The exact chirp configuration (Table~\ref{tab:chirp_config}) follows Texas Instruments' reference cascade-imaging configuration~\cite{ti_mmwcas} and is stored verbatim in each capture's \nolinkurl{*.mmwave.json} file. The DCA1000EVM cards stream interleaved complex \texttt{int16} ADC samples to TI's standard \nolinkurl{*_data.bin} and \nolinkurl{*_idx.bin} files, with one master and three slave devices per session. Released range--Doppler maps use standard FMCW processing~\cite{richards2014} with Hann windows~\cite{harris1978}; analogous pipelines produce the range--azimuth and micro-Doppler products. The benchmark uses only RD and raw IQ.

\begin{table}[tbp]
\centering
\caption{MMWCAS-RF-EVM chirp and frame configuration used for every Rad-R capture.}
\label{tab:chirp_config}
\footnotesize
\setlength{\tabcolsep}{4pt}
\begin{tabular}{lr@{\hskip 1.2em}lr}
\toprule
\textbf{Parameter} & \textbf{Value} & \textbf{Parameter} & \textbf{Value} \\
\midrule
Start freq.        & 77\,GHz       & Loops/frame ($M$)   & 64 \\
Freq.\ slope       & 78.99\,MHz/$\mu$s & Frame period    & 100\,ms \\
ADC samples ($N$)  & 256           & Cascade devices     & 4 \\
ADC sample rate    & 8\,Msps       & Total Rx           & 16 \\
TDM-MIMO Tx        & 12            & Virtual channels    & \textbf{192} \\
Eff.\ bandwidth    & 2.49\,GHz     & Range res.\        & 0.060\,m \\
Max range          & 15.2\,m       & Active Rx/device    & 4 \\
\bottomrule
\end{tabular}
\end{table}

\section{Physical Fault Induction}
\label{app:induction}

\begin{figure*}[tp]
  \centering
  \includegraphics[width=\textwidth]{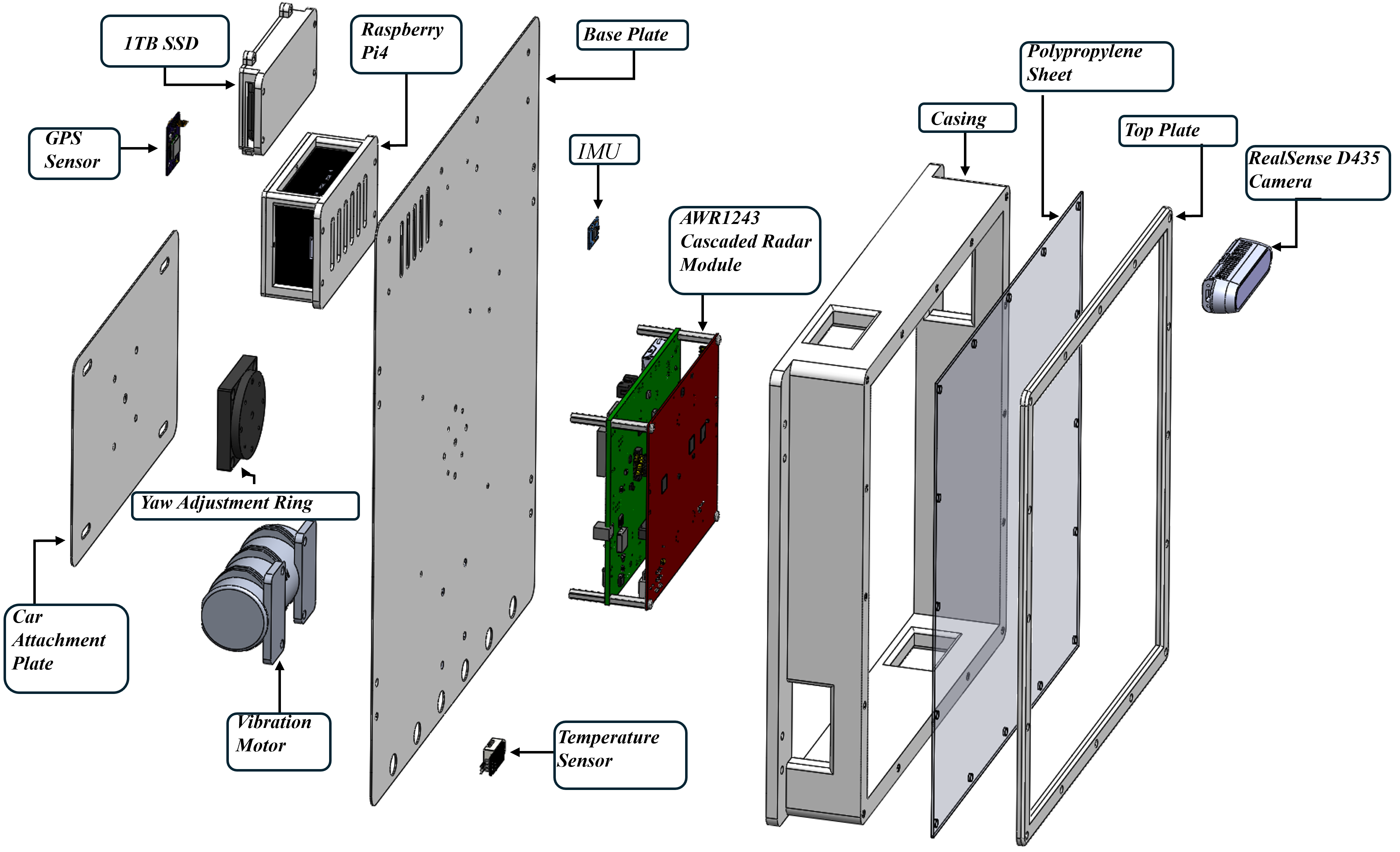}
  \caption{\textbf{The Rad-R sensor rig (exploded view).} A 4-chip TI MMWCAS-RF-EVM cascade radar (12\,Tx$\,\times\,$16\,Rx, 77\,GHz) sits behind an always-installed $1/16$-inch polypropylene radome, rigidly co-mounted with a Bosch BNO055 IMU, two DHT22 temperature probes, a u-blox NEO-M9N GPS, and an Intel RealSense D435 camera; a Raspberry Pi and 1\,TB SSD log every stream to synchronised HDF5. A yaw-adjustment ring, an eccentric-mass vibration motor, and a removable radome panel are the misalignment, vibration, and blockage fixtures, respectively. All companion sensors are time-aligned to the radar frame index.}
  \label{fig:rig}
\end{figure*}

\begin{figure*}[tp]
\centering
\includegraphics[width=\linewidth]{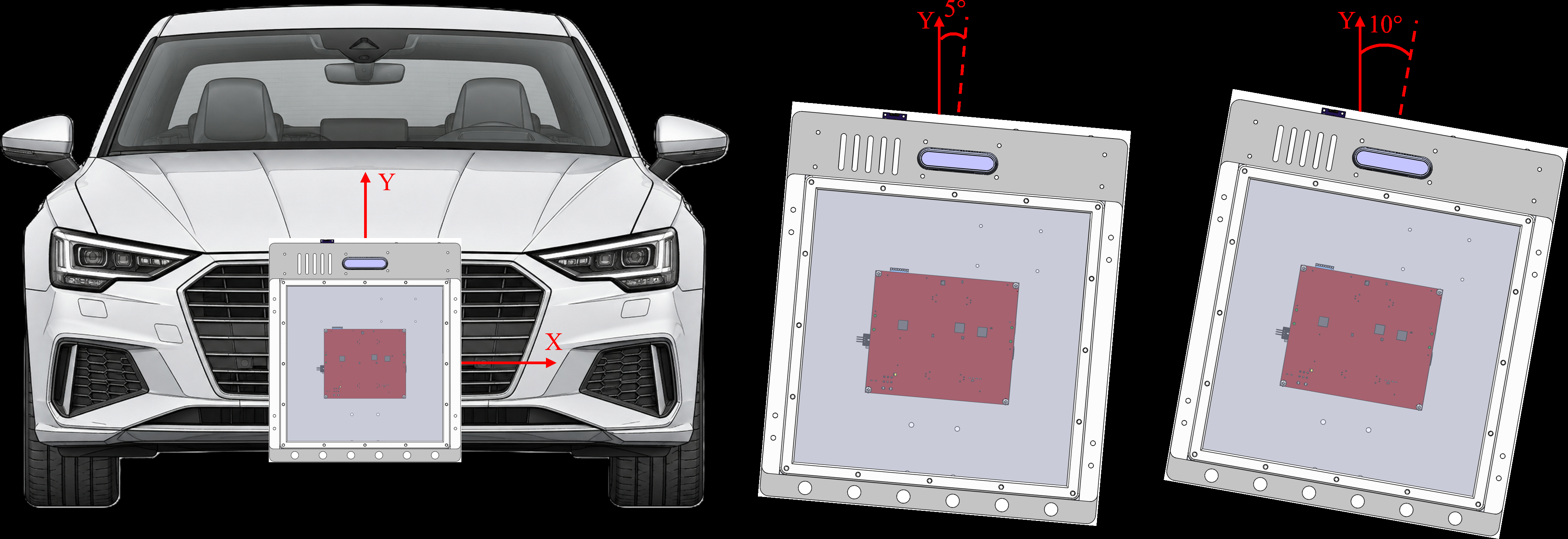}\\[3pt]
\includegraphics[width=\linewidth]{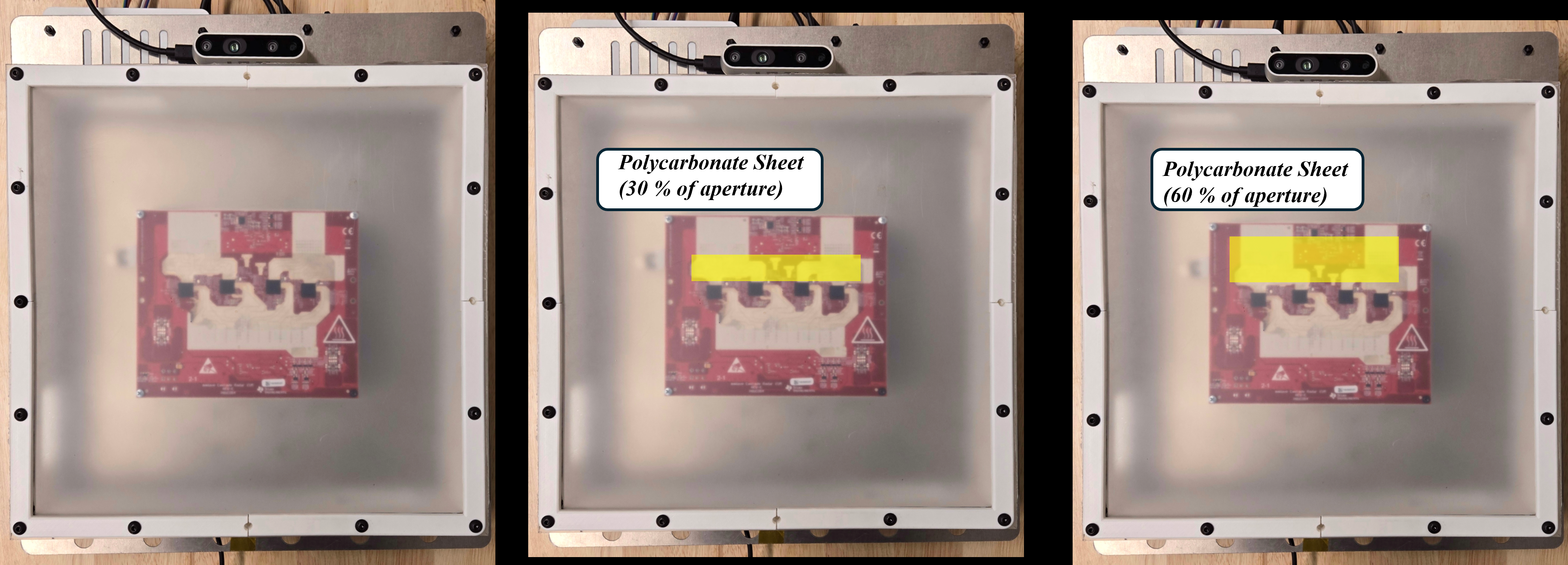}\\[3pt]
\includegraphics[width=\linewidth]{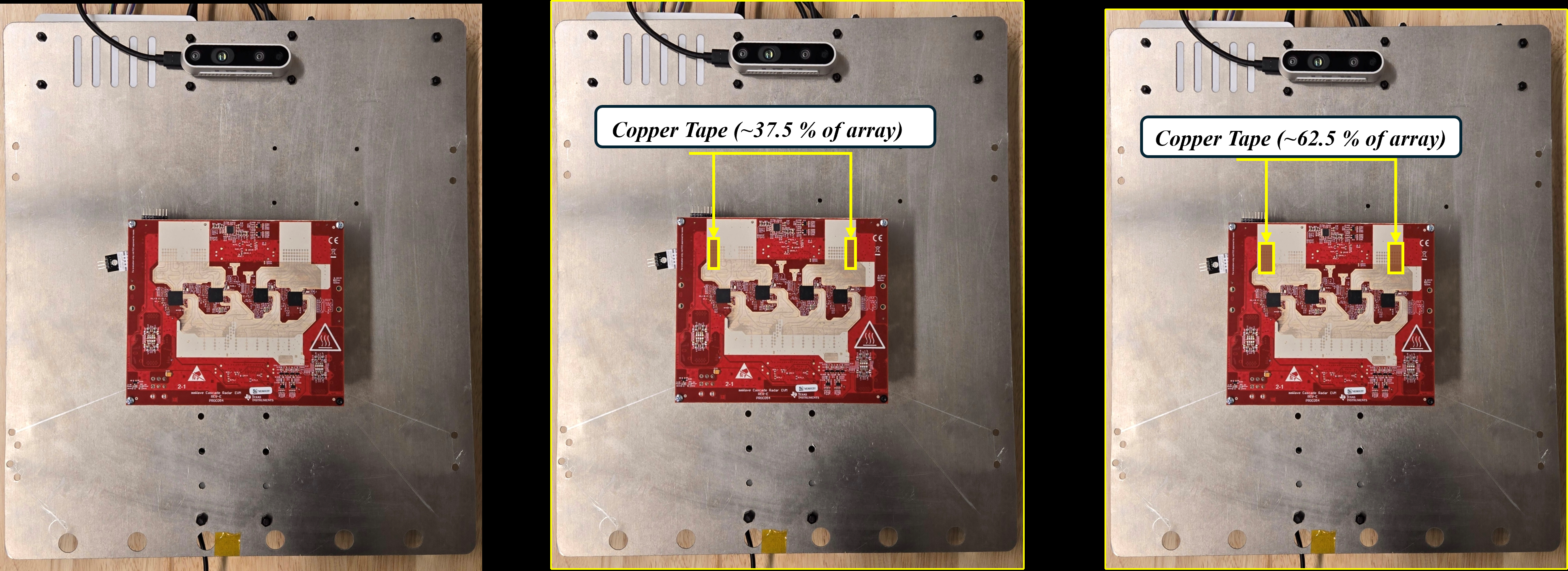}
\caption{\textbf{Physical fault induction.} \textbf{Top:} misalignment, the radar enclosure is yawed on a precision jig by $5^\circ$ then $10^\circ$ (verified with a digital inclinometer). \textbf{Middle:} blockage, a clear polycarbonate sheet covers $30\%$ then $60\%$ of the radar aperture (calibrated against a corner-reflector dB drop). \textbf{Bottom:} receive-channel degradation, copper foil tape disables $6/16$ then $10/16$ of the Rx antennas (confirmed by a single-tone calibration sweep). The leftmost panel in each row is the healthy baseline. Vibration is omitted here because its effect is dynamic rather than static; it is described in the text below.}
\label{fig:fault_induction}
\end{figure*}

Figure~\ref{fig:rig} shows the sensor rig in exploded view, including the misalignment, vibration and blockage fixtures. Each fault is produced by a physical fixture on the real radar, and its severity is set against an independent instrument rather than a nominal dial. Figure~\ref{fig:fault_induction} shows the three fixtures whose effect is directly visible on the hardware, each at the healthy, mild (S1) and severe (S2) settings.

Vibration is the one fault that leaves no before-and-after photograph, because its signature is mechanical motion rather than a fixed obstruction. An eccentric-mass DC motor is bolted to the radar bracket and shakes the sensor while it streams. Severity is set by the drive frequency, $20$\,Hz for the mild setting (S1) and $40$\,Hz for the severe setting (S2), and is verified against the onboard BNO055 IMU rather than a dial: the IMU records ground-truth tri-axial acceleration, and a capture is accepted only when its RMS acceleration falls in the band recorded for the target setting. The disturbance therefore appears in the IMU and radar micro-Doppler traces (Appendix~\ref{app:crossmodal} quantifies this agreement) but not in a still image of the hardware, which is why it has no panel in Figure~\ref{fig:fault_induction}.

\section{Sample Radar Visualizations}
\label{app:visualizations}

\begin{figure*}[tp]
  \centering
  \includegraphics[width=\linewidth]{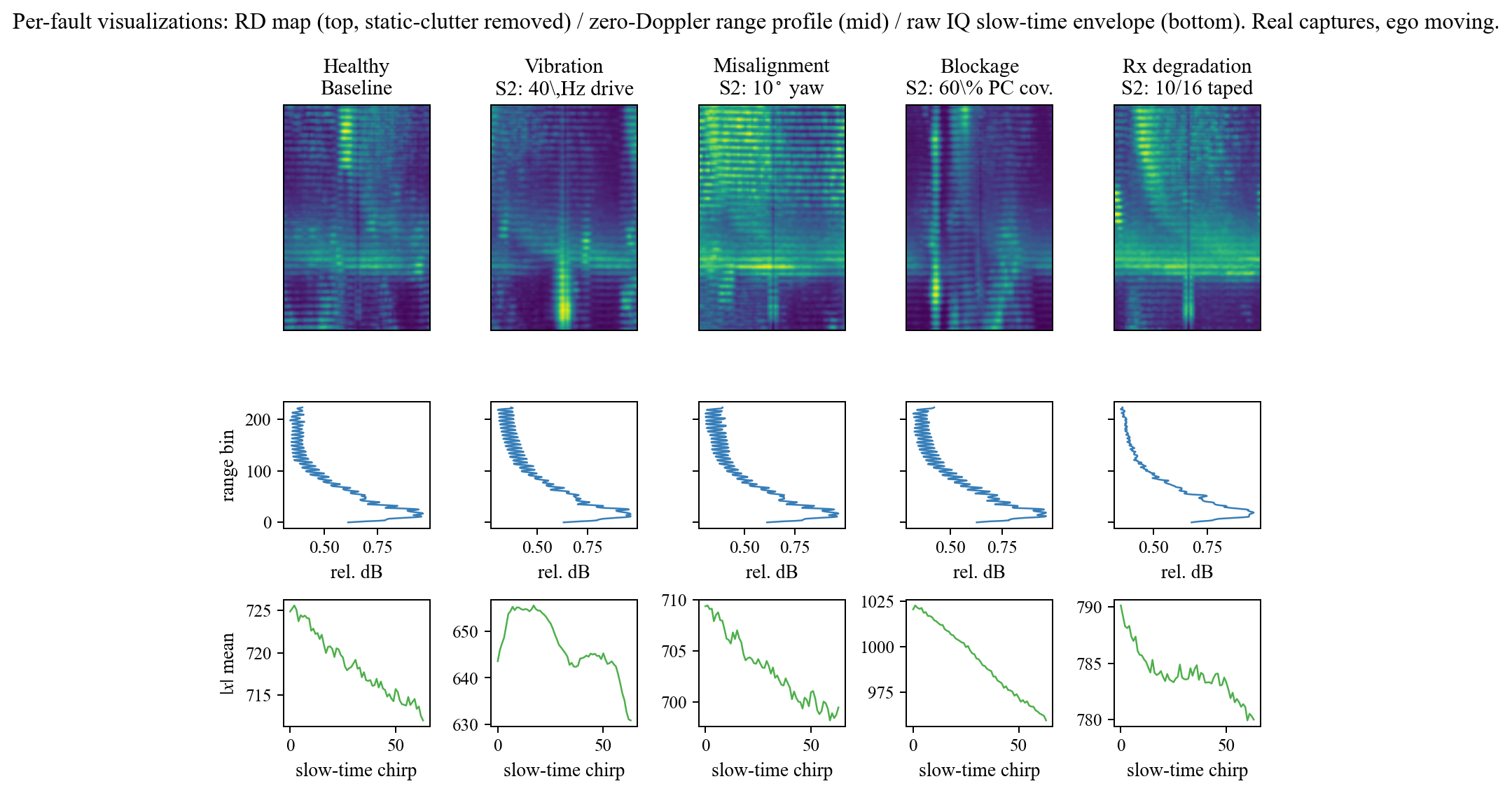}
  \caption{Expanded radar visualisation for each of the five captured classes (rows). From left: the range--Doppler map (static-clutter cancellation applied only for display), the zero-Doppler range profile, and the slow-time envelope of the raw-IQ tensor.}
  \label{fig:fault_gallery}
\end{figure*}

\begin{figure}[tp]
  \centering
  \includegraphics[width=\linewidth]{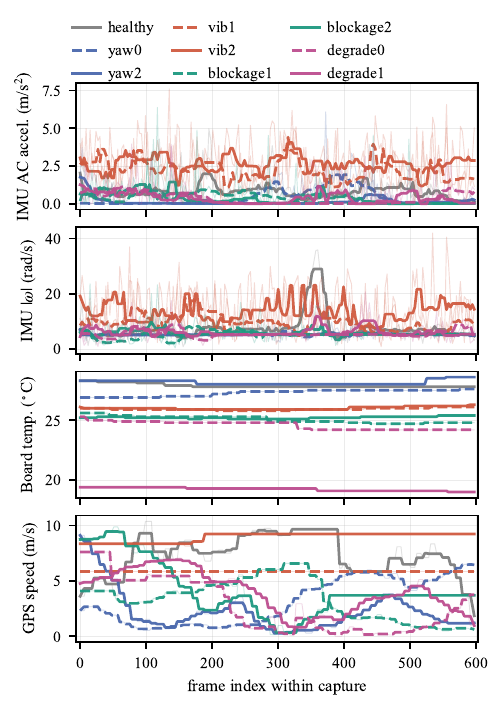}
  \caption{Synchronised companion-sensor traces, aligned to the radar frame index; each curve is one of the nine capture runs (thick: 9-frame rolling median; thin: raw). Panels: IMU AC acceleration, IMU gyro magnitude, board temperature, and GPS speed.}
  \label{fig:sensor_traces}
\end{figure}

\begin{figure*}[tp]
  \centering
  \includegraphics[width=\linewidth]{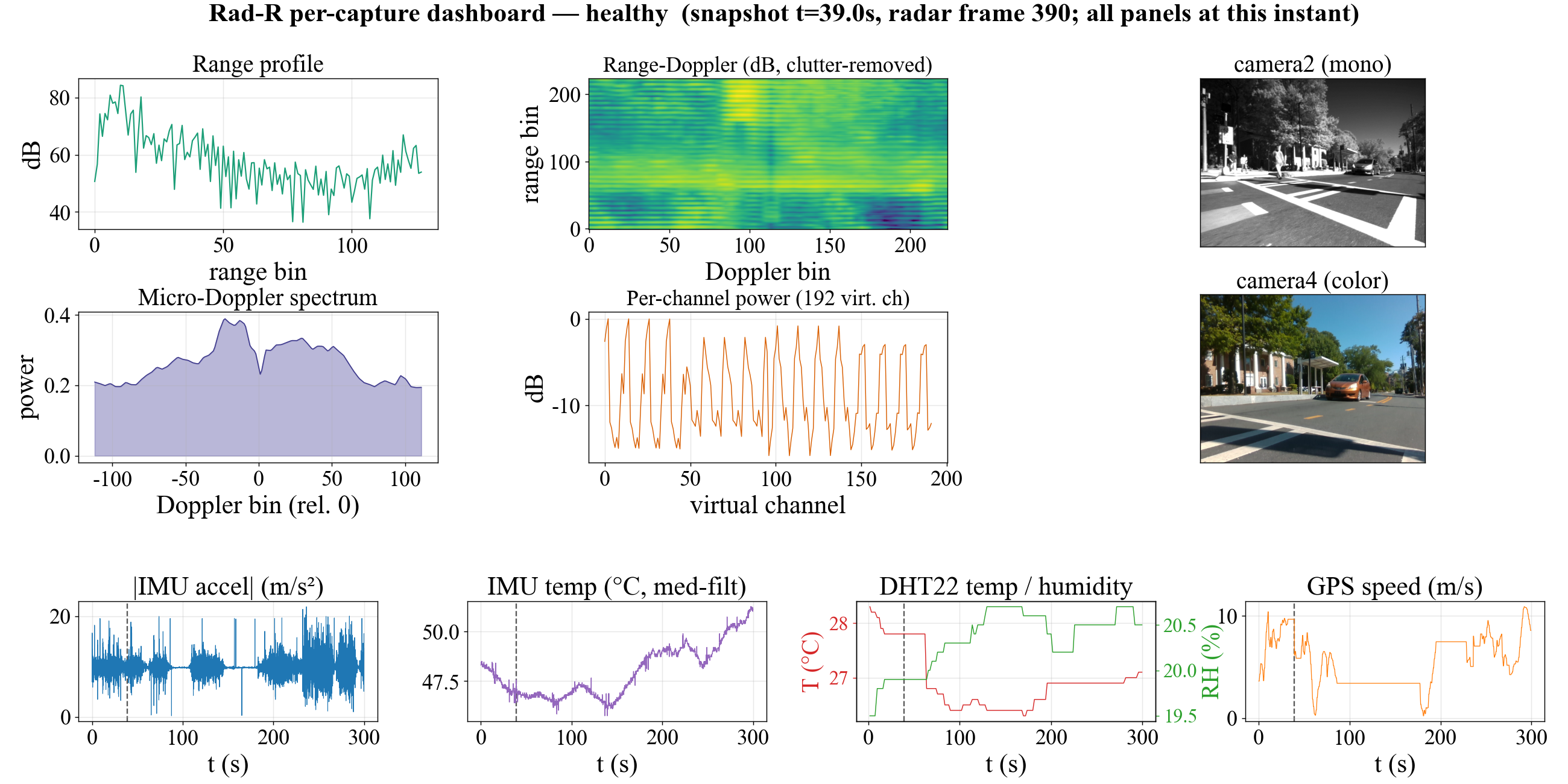}
  \caption{Per-capture dashboard for \texttt{healthy}.}
  \label{fig:dashboard_first}
\end{figure*}

\begin{figure*}[tp]
  \centering
  \includegraphics[width=\linewidth]{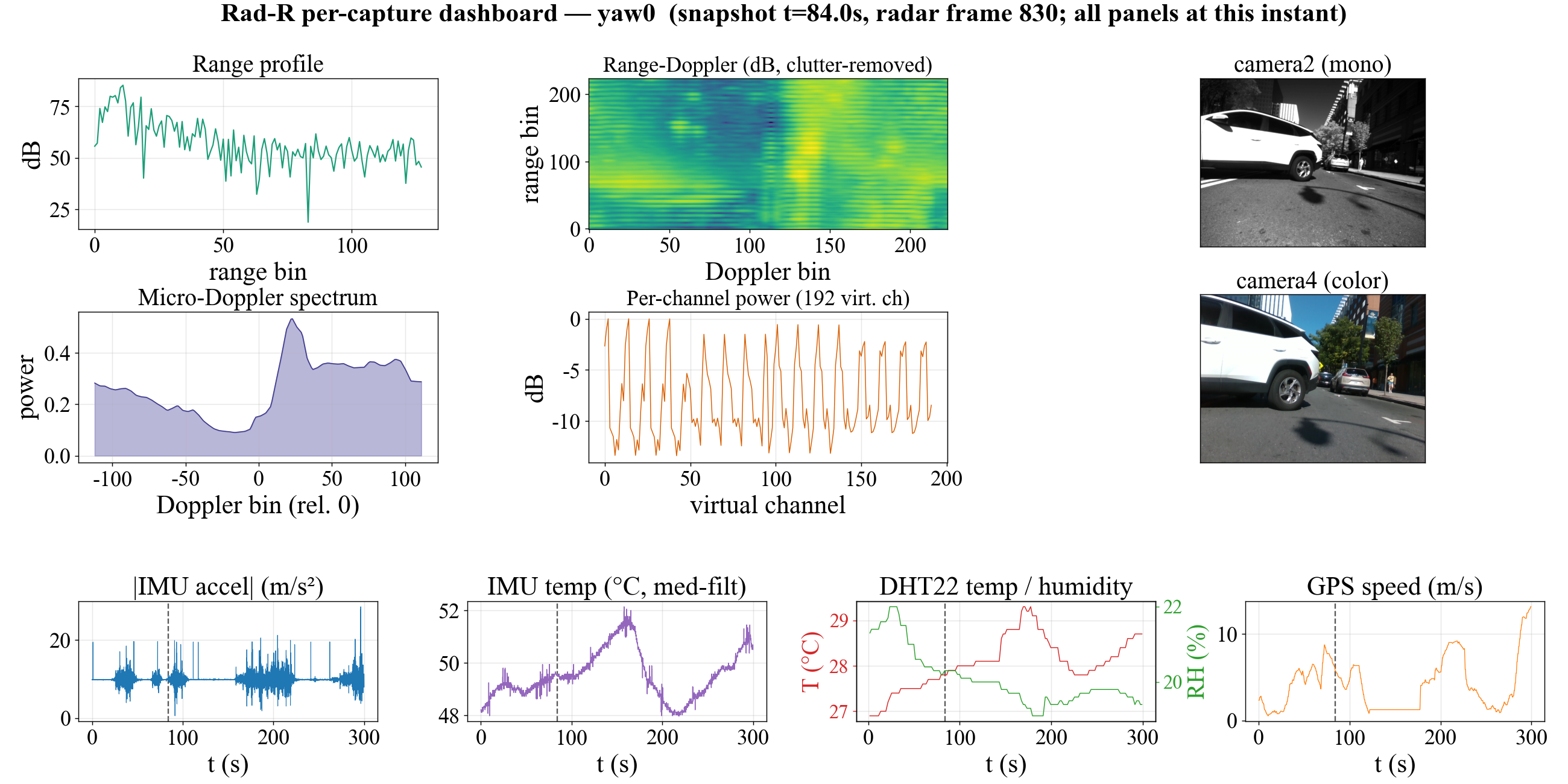}
  \caption{Per-capture dashboard for \texttt{yaw0} (5$^\circ$ misalignment).}
  \label{fig:dashboard_yaw0}
\end{figure*}

\begin{figure*}[tp]
  \centering
  \includegraphics[width=\linewidth]{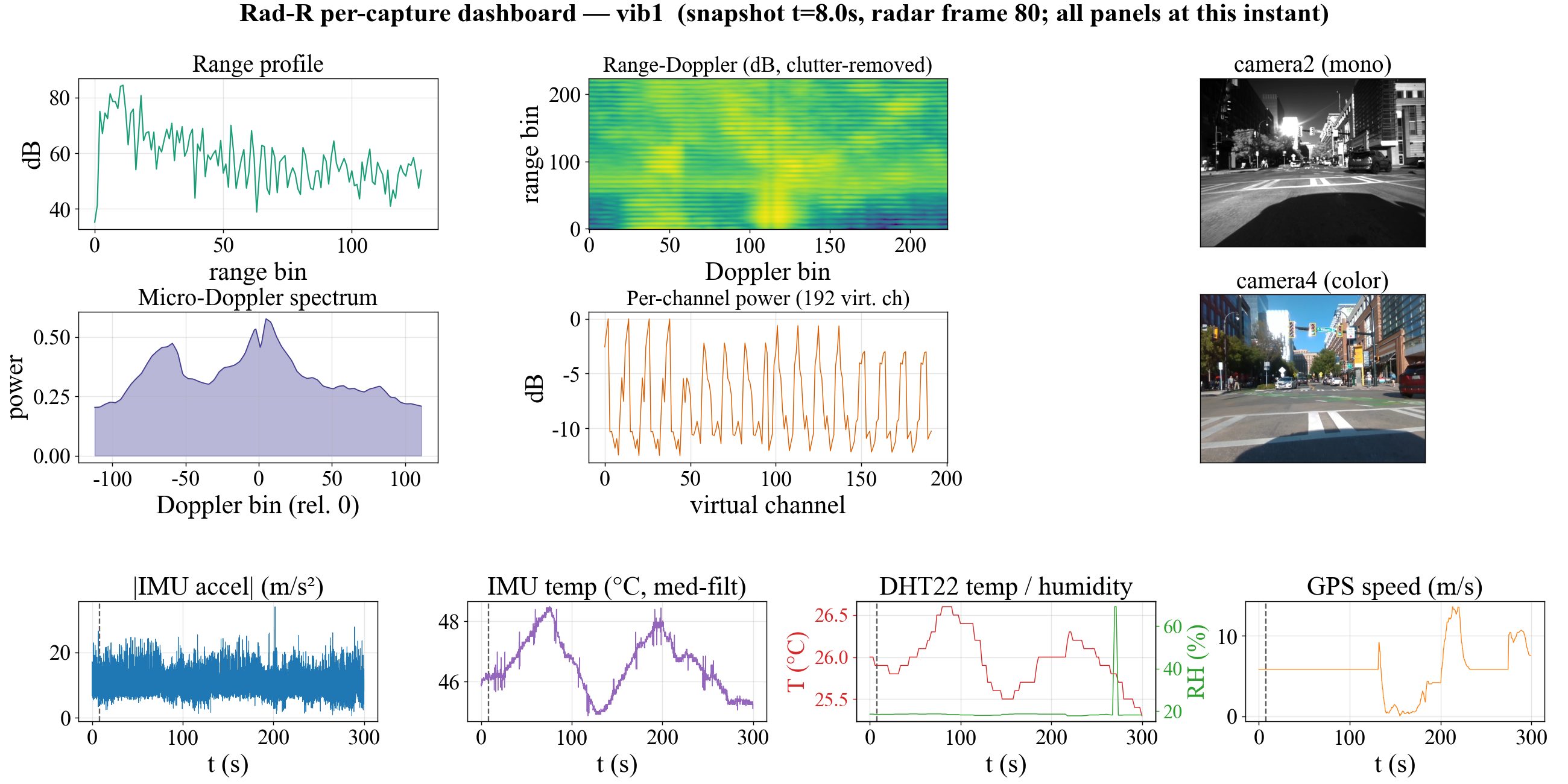}
  \caption{Per-capture dashboard for \texttt{vib1} (low-amplitude mechanical vibration).}
  \label{fig:dashboard_vib1}
\end{figure*}

\begin{figure*}[tp]
  \centering
  \includegraphics[width=\linewidth]{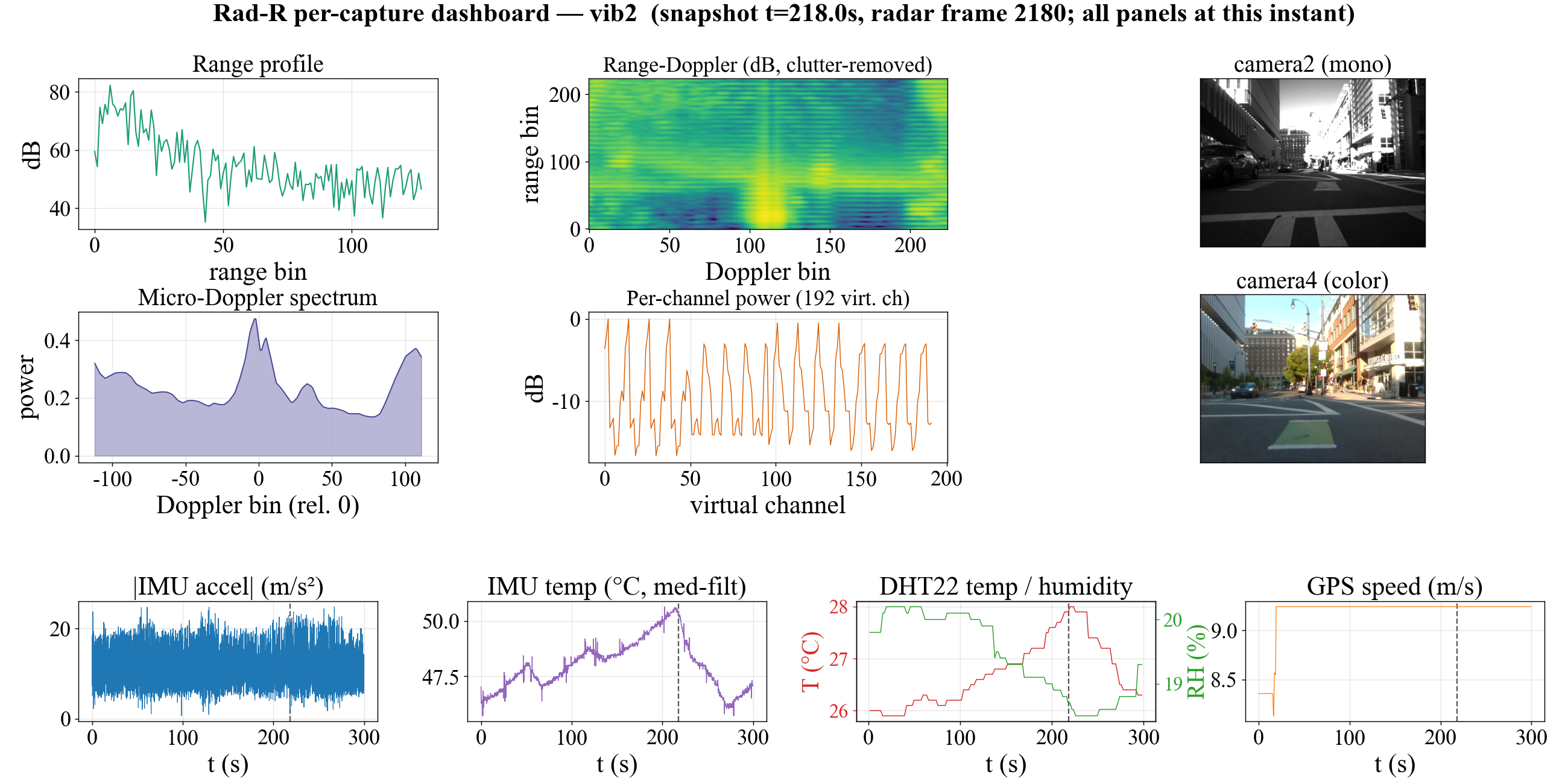}
  \caption{Per-capture dashboard for \texttt{vib2} (high-amplitude mechanical vibration).}
  \label{fig:dashboard_vib2}
\end{figure*}

\begin{figure*}[tp]
  \centering
  \includegraphics[width=\linewidth]{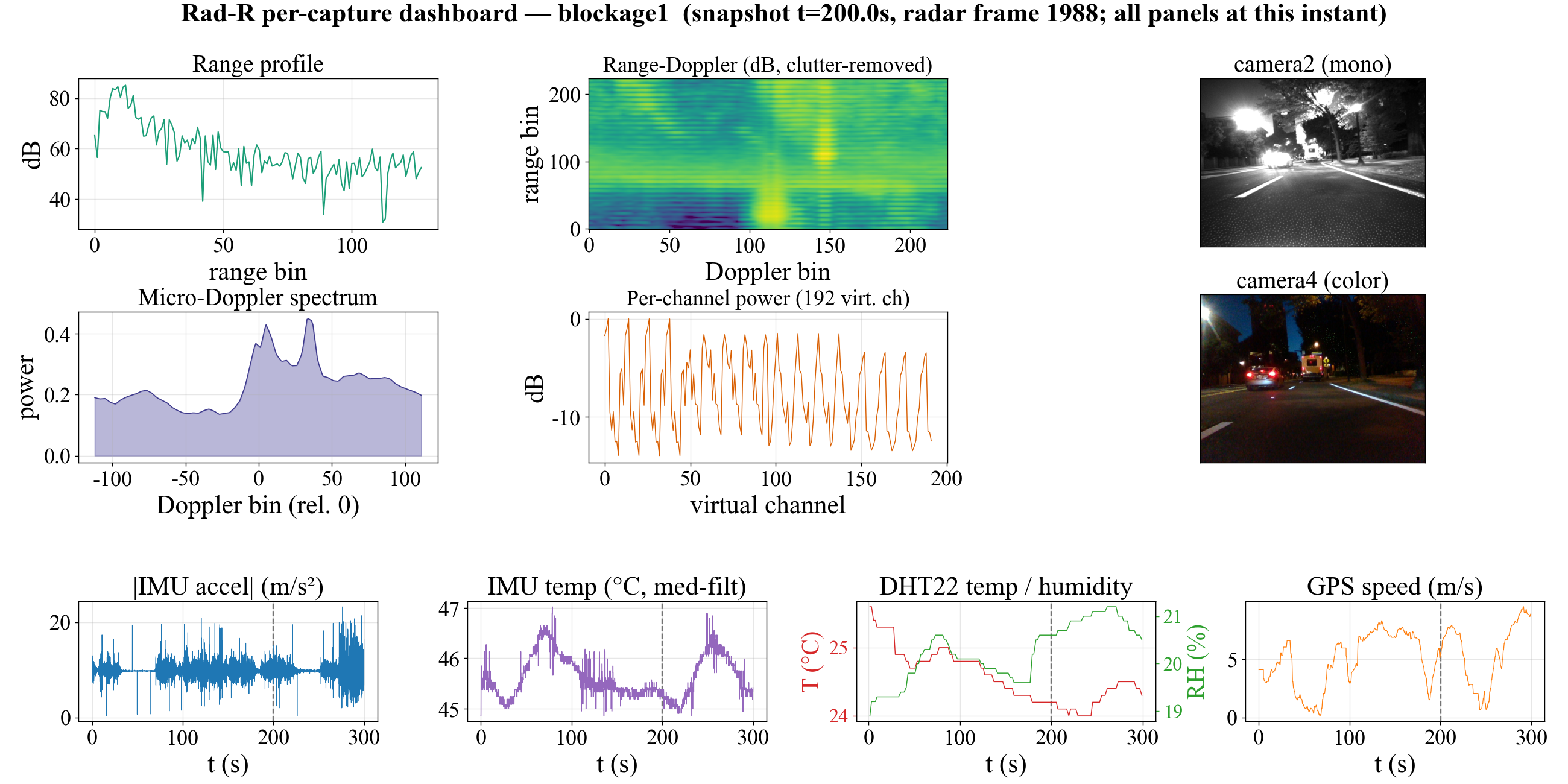}
  \caption{Per-capture dashboard for \texttt{blockage1} (S1: 30\% radome coverage).}
  \label{fig:dashboard_blockage1}
\end{figure*}

\begin{figure*}[tp]
  \centering
  \includegraphics[width=\linewidth]{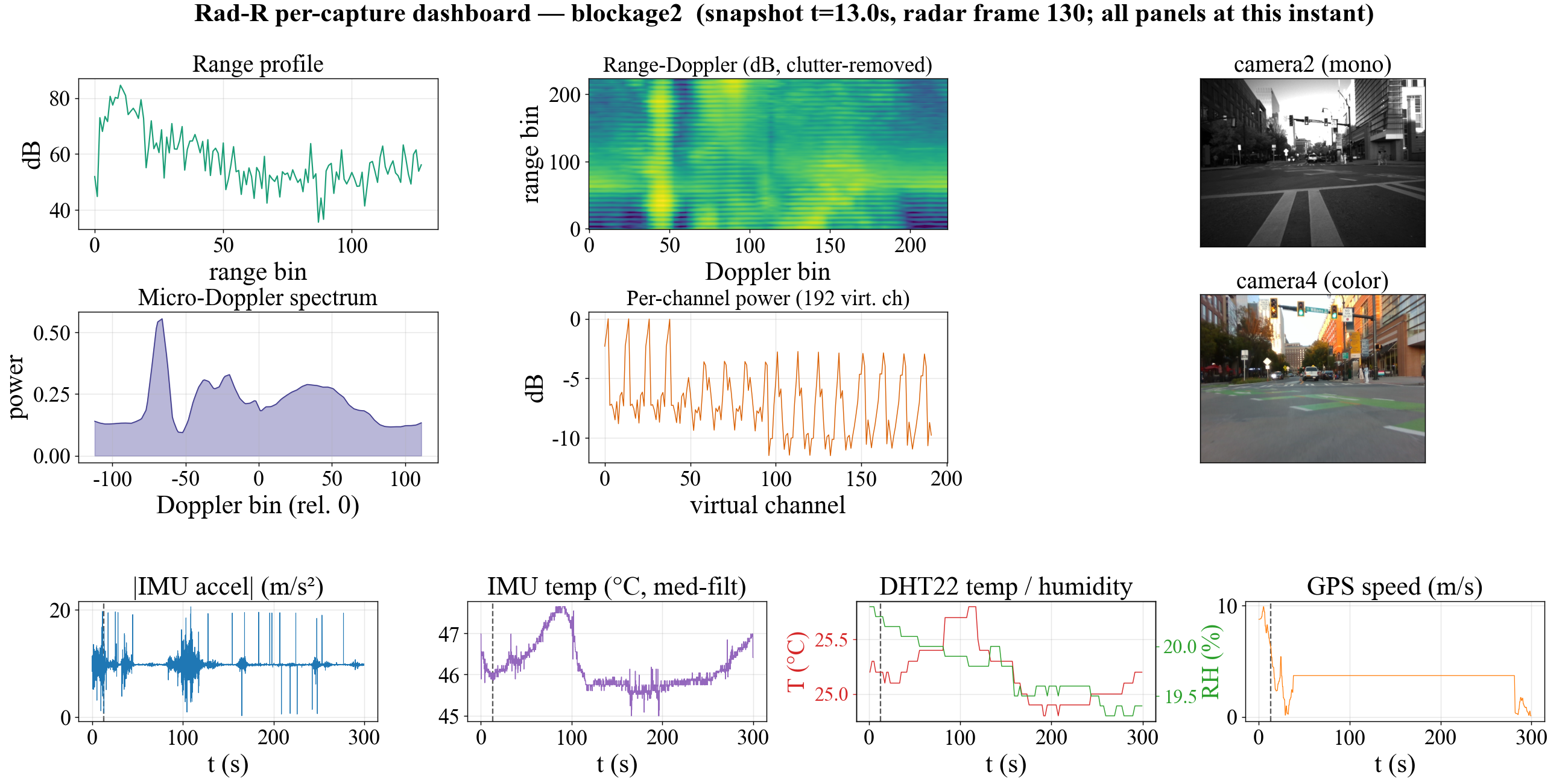}
  \caption{Per-capture dashboard for \texttt{blockage2} (S2: 60\% radome coverage).}
  \label{fig:dashboard_blockage2}
\end{figure*}

\begin{figure*}[tp]
  \centering
  \includegraphics[width=\linewidth]{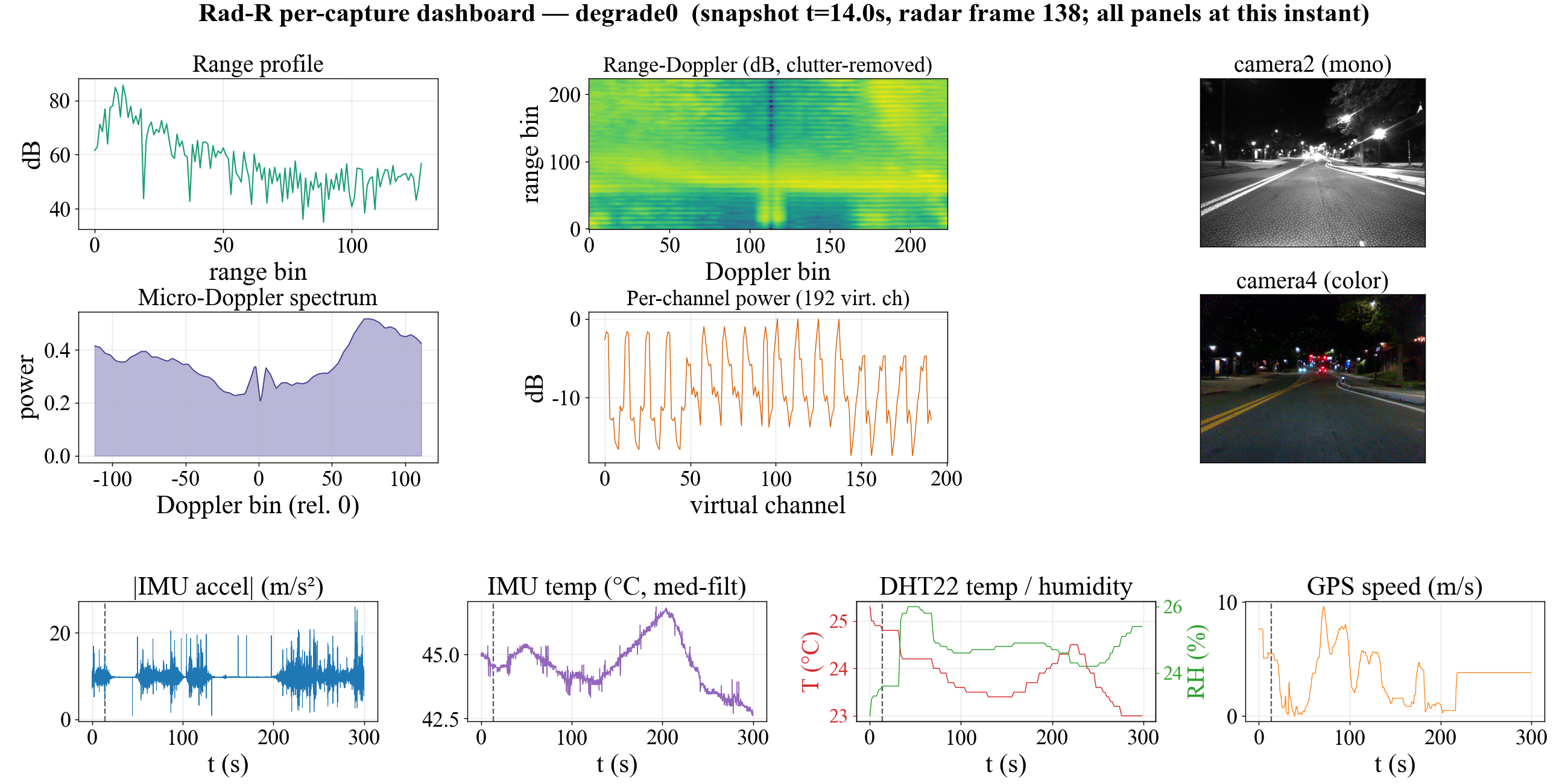}
  \caption{Per-capture dashboard for \texttt{degrade0} (S1: 6/16 Rx degraded).}
  \label{fig:dashboard_degrade0}
\end{figure*}

\begin{figure*}[tp]
  \centering
  \includegraphics[width=\linewidth]{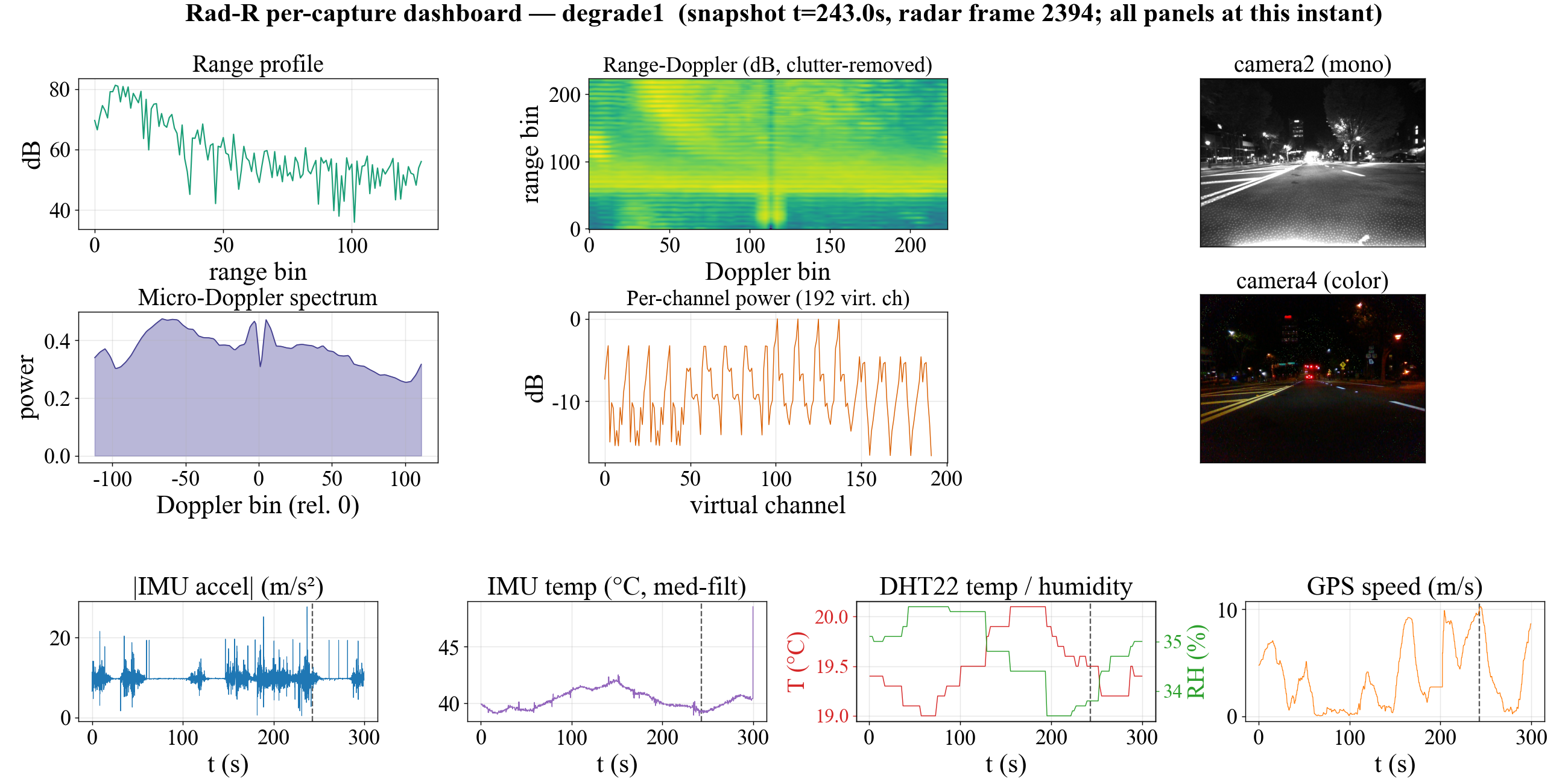}
  \caption{Per-capture dashboard for \texttt{degrade1} (S2: 10/16 Rx degraded).}
  \label{fig:dashboard_last}
\end{figure*}

The main paper presents the per-fault range--Doppler grid and one per-capture dashboard (\texttt{yaw2}). Here we add an expanded per-fault view (Fig.~\ref{fig:fault_gallery}), the synchronised companion-sensor traces (Fig.~\ref{fig:sensor_traces}), and per-capture dashboards for the remaining eight recordings (Figs.~\ref{fig:dashboard_first}--\ref{fig:dashboard_last}). Each dashboard combines a range profile, range--Doppler map, micro-Doppler spectrum, per-channel power, two synchronised camera streams, and an IMU/temperature/GPS strip.

\subsection{Per-Capture Dashboards}
\label{app:dashboards}

For each capture (the \texttt{yaw2} dashboard is in the main text), we include a multi-sensor dashboard generated from the raw ADC and ROS2 bag data. The upper-left 2$\times$2 block contains the chirp-mean range profile, a display-normalised range--Doppler map, the Doppler power profile, and power across the 192 virtual channels. The upper-right block shows the monochrome and colour camera streams at the radar timestamp marked on the lower sensor strip; faces and licence plates are Gaussian-blurred. The lower strip contains IMU acceleration magnitude, IMU temperature, DHT22 temperature and humidity, and GPS speed over the radar capture window. Vibration elevates the IMU trace, blockage depresses the near-range return, and degraded receivers create dips in per-channel power. Static misalignment is less visible in these power views because its principal signature lies in azimuth and channel phase.

\section{Datasheet for Rad-R}
\label{app:datasheet}

Following Gebru et~al.~\cite{gebru2021datasheets} and data-card guidance~\cite{datacards2022}, we provide a datasheet for Rad-R. The released Croissant 1.0 metadata encodes the corresponding machine-readable fields (Section~\ref{app:croissant}).

\subsection*{Motivation}

\paragraph{For what purpose was the dataset created?}
Rad-R was created to study how automotive-radar hardware faults propagate through the signal-processing chain into learned models, and to support fault-aware radar perception. Existing datasets either provide raw ADC under nominal operation or processed representations across varied scenes and weather; none pairs raw ADC with controlled hardware-fault conditions and independently measured severities.

\paragraph{Who created the dataset and on behalf of which entity?}
The dataset was created by the authors of this paper (affiliations on the title page).

\paragraph{Who funded the creation of the dataset?}
This work was supported by the Korea Evaluation Institute of Industrial Technology (KEIT) grant funded by the Ministry of Trade, Industry \& Energy (MOTIE) of the Republic of Korea (No.~RS-2024-00448797). The funder had no editorial influence over the dataset content or the reported conclusions.

\subsection*{Composition}

\paragraph{What do the instances that comprise the dataset represent?}
Each instance is one radar frame ($\sim$100\,ms wall-clock) containing the raw complex IQ tensor from the 4-chip cascade radar, derived range--Doppler map, and time-aligned readings from each companion sensor stream at the moment of that radar frame. Frames are grouped into 9 capture runs of 5 minutes each ($\sim$3{,}000 frames per run).

\paragraph{How many instances are there in total?}
Rad-R contains 9 capture runs covering healthy + 4 fault types $\times$ 2 severity levels each. The training cache subsamples 200 frames per capture (1{,}800 frames total). The complete raw archive comprises $\sim$26{,}800 frames. A compact representative data sample accompanies the code release, and the complete raw archive will be released in a persistent public repository.

\paragraph{Does the dataset contain all possible instances or is it a sample?}
It is a sample: one representative capture per (fault, severity) cell, in a single outdoor / lab scene with one vehicle.

\paragraph{What data does each instance consist of?}
Each radar frame contains (1) raw complex IQ with $M=64$ chirps, $N=256$ ADC samples, $R=16$ physical receivers, and $C_{\mathrm{tx}}=12$ time-multiplexed transmitters in TI's interleaved \texttt{int16} layout; (2) a pre-computed 224$\times$224, dB-scaled range--Doppler map; and (3) indices and timestamp offsets for the synchronised IMU, DHT22, GPS, and dual-camera streams.

\paragraph{Is there a label or target associated with each instance?}
Yes: a fault class label (5-way: healthy, vibration, misalignment, blockage, Rx degradation) and a severity label (3-way: S0 healthy, S1 mild, S2 severe). Labels are determined by the physical fault induction protocol, not by post-hoc human annotation.

\paragraph{Is any information missing from individual instances?}
No required radar fields are intentionally omitted. After TDM-MIMO unmixing, the cache stores both a single-Tx 16-channel slab and the full 192-channel virtual-array slab so that experiments can use either representation. Companion-sensor availability and timestamp offsets are recorded per frame.

\paragraph{Does the dataset contain data that might be considered confidential or personally identifiable?}
The two co-recorded cameras (camera2 monochrome, camera4 BGR color) may incidentally capture pedestrians or license plates during outdoor captures. \emph{Every released camera frame (camera2 and camera4) has been processed through our automated PII-redaction pipeline that detects and Gaussian-blurs human faces (using a YOLO-based face detector, $\sigma$=21\,px on the bounding box) and vehicle license plates (license-plate detector, same blur kernel) before upload}. The redacted (\texttt{*\_image\_raw\_sensored}) directories are the ones referenced from the synced HDF5; the unredacted originals never leave the authors' workstation. We further manually spot-checked a uniform random sample of 5\% of frames per capture to verify no detector misses slipped through, and re-blurred a small handful of edge cases (partial occlusions, oblique viewing angles) by hand.

\subsection*{Collection Process}

\paragraph{How was the data acquired?}
mmWave Studio captured raw \nolinkurl{*_data.bin} files via the four DCA1000EVM cards. A Raspberry Pi 4 ran \nolinkurl{collect_sensors.py} to log the BNO055 IMU at $\sim$33\,Hz, two DHT22 probes at $\sim$1\,Hz, the NEO-M9N GPS, and two cameras (monochrome and BGR colour) at $\sim$30\,fps each. Clocks were NTP-synchronised before each session and aligned at radar-frame level during post-processing.

\paragraph{What mechanisms or procedures were used to collect the data?}
Each session followed a fixed protocol: power-on and 10-minute thermal stabilization; healthy reference capture; controlled fault-induction sequence (one fault per run, severity held constant for the 5-minute capture); healthy reference recapture; per-clip metadata logging; per-file SHA-256 verification.

\paragraph{Who was involved in the data collection process?}
Two trained operators from the authors' research group. The code release includes the fault-induction SOP (\nolinkurl{docs/FAULT_INDUCTION_SOP.md}) and printable session checklist (\nolinkurl{docs/FAULT_INDUCTION_CHECKLIST.pdf}).

\paragraph{Over what timeframe was the data collected?}
Rad-R was collected on a single day in May 2026. Future releases will broaden the day/device/scene coverage.

\paragraph{Were any ethical review processes conducted?}
No human subjects research; no IRB review required. Lab safety review was conducted for the radar emission setup, the resistive heater (planned for future releases), and the secondary radar interferer (planned for future releases).

\subsection*{Preprocessing, Cleaning, and Labeling}

\paragraph{Was any preprocessing/cleaning/labeling of the data done?}
Raw ADC is stored without preprocessing. The released training cache contains derived range--Doppler maps and the raw IQ slab, both produced by \nolinkurl{scripts/build_training_cache.py}. Labels are physical-protocol labels, not annotation labels.

\paragraph{Was the ``raw'' data saved in addition to the preprocessed data?}
Yes. The raw mmWave Studio output is retained for every capture in TI's exact \texttt{*\_data.bin} format. Because a single raw capture is $\sim$37\,GB, the code release includes compact representative raw-IQ excerpts together with the loader and preprocessing code; the complete raw archive will be released in a persistent public repository.

\subsection*{Uses}

\paragraph{What tasks has the dataset been used for?}
In this paper: within-clip fault classification, severity classification, chirp-wise anytime inference, few-shot cross-capture adaptation, and confusion-matrix analysis. Additional supported tasks include severity regression from the per-capture sensor measurements, fault-aware sensor fusion, raw-IQ self-supervised pretraining, and signal-processing-pipeline benchmarking.

\paragraph{Are there tasks for which the dataset should not be used?}
Rad-R should \emph{not} be used to train safety-critical production radar-monitoring systems without extensive in-field validation. The single-day single-device capture scope is too narrow for production deployment.

\subsection*{Distribution and Maintenance}

\paragraph{How is the dataset distributed?}
The code release contains training and evaluation code, result JSONs, Croissant metadata, and a compact PII-redacted data sample. The complete $\sim$324\,GB raw archive will be released in a persistent public repository under CC BY 4.0. The code license is MIT, with the \texttt{matlab\_to\_python/} subdirectory under BSD-3-Clause to preserve TI's source license.

\paragraph{Will the dataset be updated?}
Yes. Future releases aim to broaden Rad-R along several axes (additional capture days, devices and scenes, further fault and interference modalities, and additional severity levels). Such updates will use tagged, persistent public releases with versioned identifiers.

\section{Croissant 1.0 Metadata and Responsible-AI Fields}
\label{app:croissant}

The code release includes Croissant 1.0 metadata~\cite{croissant2024} in \nolinkurl{croissant.json}. It covers file objects, file sets, record sets, and field types, plus the minimal Responsible-AI block~\cite{datacards2022}: collection and annotation protocols, release maintenance, personal information, limitations, biases, intended uses, social impact, data sources, provenance, and preprocessing. The file passes the official \texttt{mlcroissant} 1.0 validator with no schema errors; two non-blocking \texttt{@context} warnings concern upstream extension keys.

\section{Hyperparameters and Training Setup}
\label{app:hyperparams}
\label{app:hyper}

Table~\ref{tab:hp} summarizes the training hyperparameters used for every model in the benchmark. We use the same optimizer, schedule, and batch size across all models (epoch budget as noted below) so that the comparison isolates architecture; per-model hyperparameter tuning would likely improve absolute numbers but obscure architectural conclusions.

\begin{table}[tbp]
\centering
\caption{Training hyperparameters for the benchmark. Shared across models except epoch count and seeds (noted below); only the architecture differs.}
\label{tab:hp}
\small
\resizebox{\linewidth}{!}{%
\begin{tabular}{ll}
\toprule
Optimizer                       & AdamW \\
Learning rate                   & $1 \times 10^{-4}$ \\
Weight decay                    & $1 \times 10^{-4}$ \\
LR schedule                     & Cosine, $T_{\max}=$ epoch budget \\
Gradient-norm clip              & 1.0 \\
Batch size                      & 16 (32 for single-Tx cache) \\
Epochs                          & 20 (cross-sev.: 20 or 25, per text) \\
Loss                            & CE(fault) + 0.5\,CE(severity) \\
Random seed                     & fixed (multiple for multi-seed runs) \\
Checkpoint selection            & max evaluation-split fault macro-F1 (oracle; see text) \\
Hardware                        & 1$\times$ RTX A4000 (16\,GB) \\
PyTorch                         & 2.5.1+cu118 \\
Mamba kernels                   & \texttt{mamba\_ssm} 2.3.1 (CUDA 11.8) \\
\bottomrule
\end{tabular}}
\end{table}

\noindent\textbf{Checkpoint-selection limitation.} The saved within-clip and cross-severity runs do not use a separate validation partition: after each epoch, the script evaluates the designated test split and retains the checkpoint with maximum test fault macro-F1. Their reported values are therefore oracle-selected upper bounds. The few-shot protocol has no best-epoch selection and instead uses a fixed 30-step target adaptation schedule. We disclose this distinction because capture-disjoint splitting prevents frame overlap but does not remove checkpoint-selection bias.

\noindent\textbf{RadrNet hyperparameters.} The default uses embedding dimension 256, stage depths $[2,2,2]$, state size 16, convolution width 4, expansion 2, fast-axis patch size 8, and dropout 0.3. Training-only capture-invariance augmentations are a random global phase rotation $\theta \sim \mathcal{U}(0, 2\pi)$, per-Rx amplitude scaling $\sim \mathcal{U}(0.8, 1.2)$, 5\% Rx-channel dropout, and per-frame standardisation.

\section{Within-Clip Anytime Benchmark: Full Table}
\label{app:anytime_table}

Table~\ref{tab:anytime_results} gives the complete per-budget statistics for the within-clip chirp-wise anytime benchmark plotted in the main text (left panel of the two-axis figure): fault macro-F1 mean$\pm$std over five seeds with BCa 95\% bootstrap CIs, the one-sided paired-bootstrap $p$-value, and the McNemar two-sided exact $p$-value on 1\,800 pooled samples.

\begin{table*}[tp]
\centering
\caption{Chirp-wise anytime benchmark on the within-clip 80/20 MIMO split (five seeds, 20 epochs/seed). Each cell is fault macro-F1, mean$\pm$std across seeds with the BCa 95\% bootstrap CI in brackets. $p_{\mathrm{boot}}$ is the one-sided paired-bootstrap $p$-value (resampling floor $5{\times}10^{-4}$); $p_{\mathrm{McN}}$ is the McNemar two-sided exact $p$-value on 1\,800 pooled samples. Numbers verified against \texttt{results/v1\_anytime\_summary.json}.}
\label{tab:anytime_results}
\footnotesize
\setlength{\tabcolsep}{6pt}
\resizebox{\textwidth}{!}{%
\begin{tabular}{c cc c cc}
\toprule
\multirow{2}{*}{$k$} & \multicolumn{2}{c}{\textbf{F1 mean $\pm$ std [BCa 95\% CI]}} & \multirow{2}{*}{$\Delta$F1} & \multirow{2}{*}{$p_{\mathrm{boot}}$} & \multirow{2}{*}{$p_{\mathrm{McN}}$} \\
\cmidrule(lr){2-3}
 & ResNet-18 & RadrNet-DS &  &  &  \\
\midrule
 4 & $0.153 \pm 0.104$ [$.077,.232$] & $\mathbf{0.747 \pm 0.068}$ [$.673,.787$] & $+0.594$ & $5{\times}10^{-4}$ & ${<}10^{-4}$ \\
 8 & $0.128 \pm 0.071$ [$.077,.184$] & $\mathbf{0.761 \pm 0.047}$ [$.728,.803$] & $+0.633$ & $5{\times}10^{-4}$ & ${<}10^{-4}$ \\
16 & $0.090 \pm 0.020$ [$.075,.106$] & $\mathbf{0.772 \pm 0.034}$ [$.746,.800$] & $+0.682$ & $5{\times}10^{-4}$ & ${<}10^{-4}$ \\
32 & $0.121 \pm 0.042$ [$.093,.158$] & $\mathbf{0.804 \pm 0.043}$ [$.774,.843$] & $+0.683$ & $5{\times}10^{-4}$ & ${<}10^{-4}$ \\
48 & $0.402 \pm 0.083$ [$.339,.465$] & $\mathbf{0.936 \pm 0.023}$ [$.916,.954$] & $+0.534$ & $5{\times}10^{-4}$ & ${<}10^{-4}$ \\
64 & $0.926 \pm 0.017$ [$.912,.939$] & $\mathbf{0.961 \pm 0.019}$ [$.947,.976$] & $+0.035$ & $5{\times}10^{-4}$ & $5.9{\times}10^{-5}$ \\
\bottomrule
\end{tabular}}
\end{table*}

\section{Capture-Invariant SSM (RadrNet-CI / DS-CI)}
\label{app:ci}

The main paper introduces a capture-invariant SSM that reduces the cross-severity gap by encoding per-frame-standardised magnitude and \emph{relative} rather than absolute phase. This section specifies the encoding, reports the per-fold breakdown, isolates the two encoding components and gives the per-fault recall of every model (Section~\ref{app:ci_ablation}), provides the full anytime-under-cross-severity curves, and lists the hyperparameters.

\subsection{Relative-Phase Invariant Encoding}
\label{app:ci_encoding}

Let $x[c,s,r]\in\mathbb{C}$ be the raw IQ sample at chirp $c\in\{0,\dots,C{-}1\}$, ADC index $s$, and virtual channel $r$. The original RadrNet encoding stacks $(\mathrm{Re}\,x,\ \mathrm{Im}\,x,\ |x|,\ \angle x)$ and therefore exposes the absolute phase $\angle x$. Under a capture-specific transformation, each receiver experiences a constant phase offset and gain, $x'[c,s,r]=g_r\,e^{j(\phi_0+\psi_r)}\,x[c,s,r]$, where $\phi_0$ is a global phase and $\psi_r$ is a per-channel calibration phase. Because these nuisance terms are nearly constant within a recording but can differ between recordings, absolute phase can reveal capture identity and provide a shortcut under within-clip evaluation.

RadrNet-CI removes this by encoding two invariants. First, a per-frame standardised magnitude
\[
\tilde m[c,s,r] = \frac{|x[c,s,r]| - \mu}{\sigma},
\]
where $\mu$ and $\sigma$ are the mean and standard deviation over $(c,s,r)$. This removes a frame-wide offset and scale and is exactly invariant to global positive gain; it does not remove arbitrary per-channel gain changes. Second, the \emph{relative} (slow-time) phase, the chirp-to-chirp difference
\[
\Delta\phi[c,s,r] = \angle x[c{+}1,s,r] - \angle x[c,s,r],
\]
encoded as $(\sin\Delta\phi,\ \cos\Delta\phi)$ to avoid wrap discontinuities. Differencing along $c$ cancels any term constant in $c$: with $x'$ as above, $\angle x'[c{+}1,s,r]-\angle x'[c,s,r]=\Delta\phi[c,s,r]$, so both $\phi_0$ \emph{and} $\psi_r$ vanish exactly. $\Delta\phi$ preserves the slow-time phase \emph{dynamics}; which fault signatures each component actually carries is measured in Section~\ref{app:ci_ablation}. The resulting input has $3R$ channels $(\tilde m,\sin\Delta\phi,\cos\Delta\phi)$ versus the $4R$ of RadrNet; the invariance is a property of the architecture, not of a training-time augmentation. RadrNet-DS-CI feeds this invariant branch and a small RD-CNN branch (identical to RadrNet-DS's) into the same 2-way input-dependent gate.

\subsection{Full Cross-Severity Ranking}
\label{app:crosssev_full}

Table~\ref{tab:crosssev_full} lists the per-model controlled cross-severity fault macro-F1 (three-seed mean$\pm$std) with input modality; these are the exact values behind the main-text ranking figure.

\begin{table}[tbp]
\centering
\caption{Controlled cross-severity 2-fold benchmark, fault macro-F1 (mean$\pm$std over three seeds). From \texttt{results/v1\_2fold\_*\_seed*.json}.}
\label{tab:crosssev_full}
\small
\setlength{\tabcolsep}{6pt}
\begin{tabular}{ll c}
\toprule
\textbf{Model} & \textbf{Input} & \textbf{F1} \\
\midrule
\textbf{RadrNet-DS-CI} & RD + inv.\ IQ & $\mathbf{0.663 \pm 0.009}$ \\
ResNet-18      & RD map      & $0.628 \pm 0.015$ \\
RadrNet-CI     & inv.\ IQ    & $0.612 \pm 0.015$ \\
DeiT-III       & RD map      & $0.581 \pm 0.024$ \\
Swin-T         & RD map      & $0.568 \pm 0.010$ \\
EfficientNetV2 & RD map      & $0.566 \pm 0.013$ \\
RadrNet        & raw IQ      & $0.545 \pm 0.006$ \\
PVTv2          & RD map      & $0.538 \pm 0.006$ \\
ConvNeXt-V2    & RD map      & $0.529 \pm 0.018$ \\
RadrNet-DS     & RD + raw IQ & $0.485 \pm 0.041$ \\
ViT-Small      & RD map      & $0.480 \pm 0.010$ \\
\bottomrule
\end{tabular}
\end{table}

\subsection{Per-Fold Cross-Severity Breakdown}
\label{app:ci_folds}

\begin{figure*}[tp]
  \centering
  \includegraphics[width=\linewidth]{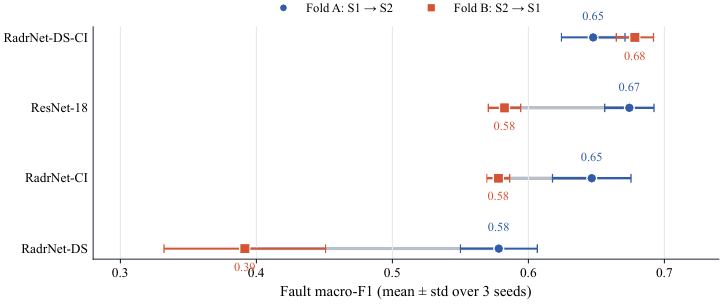}
  \caption{Per-fold controlled cross-severity fault macro-F1 (three-seed means): Fold~A (train S1, test S2) vs.\ Fold~B (train S2, test S1) for each model.}
  \label{fig:perfold}
\end{figure*}

Figure~\ref{fig:perfold} reports the two folds separately. Fold~A trains on mild (S1) and tests on severe (S2); fold~B reverses this. Fold~B is the direction in which the off-the-shelf raw-IQ models collapsed: RadrNet-DS scores only $0.392$ there. RadrNet-DS-CI recovers fold~B to $0.678$, the single largest source of its overall gain, while keeping fold~A competitive; this is the quantitative form of the ``fold-B fix'' claimed in the main text.

\subsection{Encoding Ablation and Per-Fault Breakdown}
\label{app:ci_ablation}

To isolate the two components of the capture-invariant encoding, we trained RadrNet-CI with each component alone---per-frame-standardised magnitude only ($R$ input channels) and relative phase only ($(\sin\Delta\phi,\cos\Delta\phi)$, $2R$ channels)---keeping the backbone, augmentations, recipe, 2-fold cross-severity protocol and checkpoint rule identical to the full model (three seeds each). Table~\ref{tab:ci_ablation} reports fault macro-F1 with the per-fold means and, for every cross-severity model of the main text, the per-fault recall on the \emph{unseen} test captures pooled over both folds and three seeds.

\begin{table*}[tbp]
\centering
\caption{Encoding ablation and per-fault breakdown under the controlled cross-severity 2-fold protocol (three seeds). F1: fault macro-F1, mean$\pm$std over seeds; Fold~A/B: per-fold means (A: train S1, test S2; B: train S2, test S1); Vib./Misal./Block./Rx-deg.: recall (\%) of each true fault on the unseen test captures. Encoding: abs.\ = real/imag/magnitude/absolute phase ($4R$); mag.\ = standardised magnitude only ($R$); rel.\ = relative phase only ($2R$); both = RadrNet-CI encoding ($3R$). From \texttt{results/v1\_2fold\_*.json}.}
\label{tab:ci_ablation}
\small
\setlength{\tabcolsep}{6pt}
\begin{tabular}{llccccccc}
\toprule
\textbf{Model} & \textbf{Encoding, input} & \textbf{F1} & \textbf{Fold A} & \textbf{Fold B} & \textbf{Vib.} & \textbf{Misal.} & \textbf{Block.} & \textbf{Rx-deg.} \\
\midrule
RadrNet        & abs., IQ        & $0.545 \pm 0.006$ & 0.655 & 0.434 & 84.7 & 77.2 & 23.4 & 94.4 \\
RadrNet-DS     & abs., IQ+RD     & $0.485 \pm 0.041$ & 0.578 & 0.392 & 89.2 & 74.9 & 26.2 & 54.2 \\
RadrNet-CI     & mag., IQ        & $0.631 \pm 0.028$ & 0.724 & 0.538 & 73.0 & 67.0 & 67.3 & 86.3 \\
RadrNet-CI     & rel., IQ        & $0.399 \pm 0.025$ & 0.388 & 0.410 & 30.8 & 43.4 & 49.2 & 67.7 \\
RadrNet-CI     & both, IQ        & $0.612 \pm 0.015$ & 0.647 & 0.578 & 63.0 & 53.6 & 69.8 & 95.8 \\
RadrNet-DS-CI  & both, IQ+RD     & $\mathbf{0.663 \pm 0.009}$ & 0.648 & 0.678 & 91.7 & 72.5 & 56.8 & 94.7 \\
ResNet-18      & RD map          & $0.628 \pm 0.015$ & 0.675 & 0.582 & 87.3 & 83.7 & 54.2 & 77.1 \\
\bottomrule
\end{tabular}
\end{table*}

\noindent\textbf{Which component carries the gain.} Per-frame magnitude standardisation is the dominant factor: on its own it raises the mean from $0.545$ to $0.631$ and recovers blockage recall from $23.4\%$ to $67.3\%$. Relative phase on its own is insufficient ($0.399$, below the absolute-phase baseline) and does not raise the mean when added to magnitude ($0.612$ vs.\ $0.631$, within the seed spread). Its contribution is confined to the harder severe$\to$mild fold~B ($0.538 \to 0.578$), where it lifts blockage ($76.8\% \to 89.3\%$) and Rx-degradation ($72.7\% \to 91.8\%$) recall, at a cost in fold~A ($0.724 \to 0.647$, mainly misalignment, $93.7\% \to 66.3\%$). Relative phase alone recovers vibration poorly ($30.8\%$), so the chirp-to-chirp phase difference is not where the vibration signature is most readily read off. The invariance property of Section~\ref{app:ci_encoding} (exact cancellation of constant global and per-channel phase offsets) is unaffected, but the cross-severity gain of RadrNet-CI over RadrNet should be attributed mainly to magnitude standardisation. Without checkpoint selection (fixed 20-epoch schedule, last epoch), the full-frame fault macro-F1 is $0.497$ (magnitude only), $0.382$ (relative phase only) and $0.580$ (both).

\noindent\textbf{Why the RD branch hurts the absolute-phase model but helps the invariant one.} With absolute phase, adding the RD branch collapses fold-B Rx-degradation recall from $98.8\%$ to $17.3\%$, which is the whole drop from $0.545$ to $0.485$; with the invariant encoding, the RD branch lifts fold-B vibration ($31.5\% \to 85.5\%$) and misalignment ($40.8\% \to 64.0\%$) recall at a modest blockage cost ($89.3\% \to 77.7\%$). Every model recovers every fault on captures it never saw during training (the only cell near chance is absolute-phase blockage in fold~B), so the models separate faults rather than captures; capture-specific cues instead hurt transfer.

\subsection{Anytime Under Cross-Severity}
\label{app:ci_anytime}

Table~\ref{tab:ci_anytime} gives the anytime curves under the cross-severity split. RadrNet-DS-CI is substantially weaker at partial budgets and reaches its highest prefix-sweep score only at $k{=}64$, consistent with the RD branch benefiting from a better-resolved Doppler spectrum. RadrNet-CI reaches $0.479$ fault macro-F1 from $4$ of $64$ chirps (three-seed mean) and increases monotonically, giving the strongest reported small-budget performance under cross-severity at the cost of a lower full-frame score than DS-CI. ResNet-18 shows a similar dependence on chirp budget. The absolute-phase RadrNet-DS remains below RadrNet-CI throughout and trails ResNet-18 at the full frame ($0.441$ vs. $0.591$). The DS-CI $k{=}64$ value ($0.601$) is produced by the separate prefix-evaluation path and flat averaging over six seed$\times$fold runs; the headline $0.663$ instead averages fold-level full-frame metrics within each seed and then across seeds. The endpoints are therefore not directly interchangeable. These results show a trade-off between early raw-IQ evidence and full-frame RD evidence; they do not isolate a single causal component.

\begin{table}[tbp]
\centering
\caption{Anytime-under-cross-severity fault macro-F1 at chirp budget $k$. RadrNet-CI and RadrNet-DS-CI are three-seed means over seeds$\times$folds; ResNet-18 and the absolute-phase RadrNet-DS are a single seed (mean over the 2 folds). From the \texttt{anytime\_curve} fields of \texttt{v1\_2fold\_\{dsci,ci,anytime\}\_seed*.json}.}
\label{tab:ci_anytime}
\small
\setlength{\tabcolsep}{5pt}
\resizebox{\linewidth}{!}{%
\begin{tabular}{ccccc}
\toprule
$k$ & \textbf{RadrNet-CI} & \textbf{RadrNet-DS-CI} & \textbf{ResNet-18} & \textbf{RadrNet-DS} \\
\midrule
 4 & $\mathbf{0.479}$ & $0.304$ & $0.127$ & $0.257$ \\
 8 & $\mathbf{0.488}$ & $0.308$ & $0.128$ & $0.273$ \\
16 & $\mathbf{0.504}$ & $0.310$ & $0.142$ & $0.283$ \\
32 & $\mathbf{0.546}$ & $0.319$ & $0.146$ & $0.287$ \\
48 & $\mathbf{0.571}$ & $0.332$ & $0.181$ & $0.295$ \\
64 & $0.580$ & $\mathbf{0.601}$ & $0.591$ & $0.441$ \\
\bottomrule
\end{tabular}}
\end{table}

\subsection{Few-Shot Label-Budget Axis}
\label{app:ci_fewshot}

\begin{figure}[tbp]
  \centering
  \includegraphics[width=\linewidth]{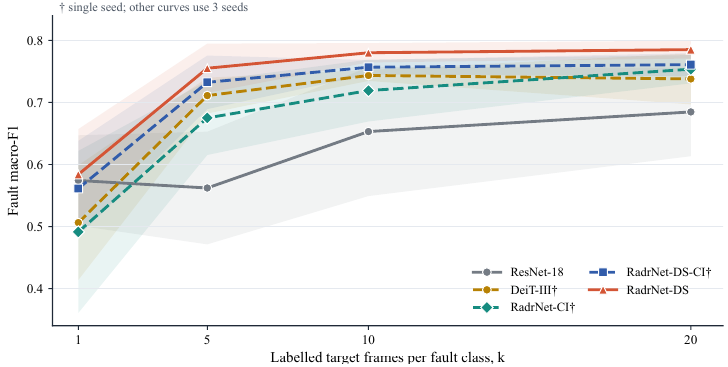}
  \caption{Few-shot label-budget curve: fault macro-F1 vs.\ labelled frames per class $k$ for the five compared models. RadrNet-DS and ResNet-18 show three-seed aggregates; the other curves are single-seed. Error bars and exact values are in Table~\ref{tab:fewshot}.}
  \label{fig:fewshot_all}
\end{figure}

Table~\ref{tab:fewshot} and Figure~\ref{fig:fewshot_all} report all five models under the identical few-shot protocol of the main text (15-epoch base training, 30 adaptation steps at lr $10^{-4}$, 2 folds $\times$ 3 support draws per seed). RadrNet-DS and ResNet-18 use three seeds; DeiT-III, RadrNet-CI, and RadrNet-DS-CI use one. RadrNet-DS-CI places second overall, above every published baseline from $k{=}5$ (e.g.\ $0.732$ vs. DeiT-III's $0.711$ at $k{=}5$), while the absolute-phase RadrNet-DS remains the most label-efficient ($0.755/0.780/0.785$ at $k{=}5/10/20$). The invariant encoding thus trades a small amount of few-shot adaptability for its cross-severity generalisation and anytime robustness, completing the trade-space picture: no single variant wins every axis.

\begin{table}[tbp]
\centering
\caption{Few-shot cross-capture adaptation (label-budget axis), fault macro-F1 at label budgets $k\geq1$. RadrNet-DS and ResNet-18 are aggregated over three seeds $\times$ two folds $\times$ three support draws; the other models use one seed $\times$ two folds $\times$ three draws.}
\label{tab:fewshot}
\small
\setlength{\tabcolsep}{4pt}
\resizebox{\linewidth}{!}{%
\begin{tabular}{c ccccc}
\toprule
$k$ & ResNet-18 & DeiT-III & RadrNet-DS & RadrNet-CI & RadrNet-DS-CI \\
\midrule
 1 & $0.574 \pm 0.073$ & $0.506 \pm 0.093$ & $\mathbf{0.584 \pm 0.073}$ & $0.491 \pm 0.131$ & $0.561 \pm 0.077$ \\
 5 & $0.562 \pm 0.091$ & $0.711 \pm 0.029$ & $\mathbf{0.755 \pm 0.040}$ & $0.675 \pm 0.060$ & $0.732 \pm 0.043$ \\
10 & $0.653 \pm 0.104$ & $0.744 \pm 0.009$ & $\mathbf{0.780 \pm 0.016}$ & $0.719 \pm 0.050$ & $0.757 \pm 0.012$ \\
20 & $0.685 \pm 0.071$ & $0.738 \pm 0.041$ & $\mathbf{0.785 \pm 0.014}$ & $0.754 \pm 0.023$ & $0.761 \pm 0.018$ \\
\bottomrule
\end{tabular}}
\end{table}

\subsection{Hyperparameters}
\label{app:ci_hyper}

RadrNet-CI and RadrNet-DS-CI reuse RadrNet's backbone and training recipe exactly: \texttt{embed\_dim=256}, \texttt{depths=[2,2,2]}, \texttt{d\_state=16}, \texttt{d\_conv=4}, \texttt{expand=2}, \texttt{patch\_size\_fast=8}, \texttt{dropout=0.3}; AdamW at lr $10^{-4}$, weight decay $10^{-4}$, cosine schedule, gradient-norm clip 1.0, batch size 16, 20 epochs, oracle checkpoint selection by evaluation-split fault macro-F1, and loss $\mathrm{CE}(\text{fault})+0.5\,\mathrm{CE}(\text{severity})$. The only architectural change is the input encoding: the invariant branch ingests the $3R$-channel $(\tilde m,\sin\Delta\phi,\cos\Delta\phi)$ tensor of Section~\ref{app:ci_encoding} in place of RadrNet's $4R$-channel real/imag/mag/abs-phase tensor; the absolute-phase augmentations are consequently unnecessary for the invariant branch. Parameter counts are $\sim$4.1\,M (RadrNet-CI) and $\sim$4.7\,M (RadrNet-DS-CI), both smaller than RadrNet-DS ($\sim$5.1\,M). The cross-severity benchmark uses the 2-fold train-one-severity/test-the-other protocol of the main text; both CI and DS-CI are run for three seeds.

\section{Cross-Modal Comparison: Radar Micro-Doppler vs.\ IMU}
\label{app:crossmodal}

We compare a hand-computed radar feature with the independently measured IMU signal. For each frame of the \texttt{healthy}, \texttt{vib1}, and \texttt{vib2} captures (200 frames per capture), we compute the off-centre Doppler energy $1-p_0$, where $p_0$ is the zero-Doppler bin's share of the slow-time power spectrum averaged over fast time and 192 virtual channels. IMU vibration energy is the RMS of the de-meaned accelerometer-plus-gyroscope magnitude over the corresponding frame window. Pooled over frames, the descriptive rank association is Spearman $\rho=0.41$: conditions with greater IMU vibration also show broader radar micro-Doppler. Within either vibration capture, however, the frame-wise association is approximately zero. Because the 600 frames come from only three condition-level captures and are temporally correlated, we do not interpret a frame-wise permutation test as capture-level significance; the result supports between-condition ordering rather than within-capture tracking.

\begin{figure}[tbp]
  \centering
  \includegraphics[width=0.9\linewidth]{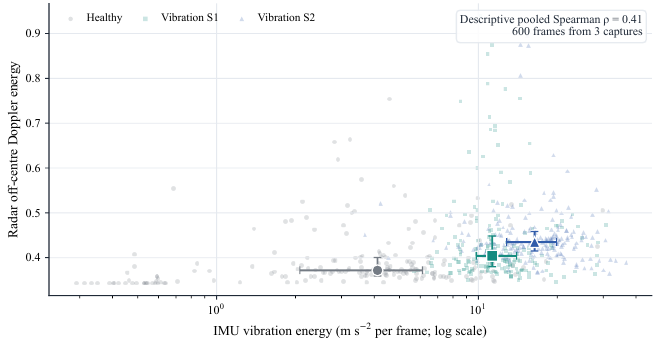}
  \caption{Descriptive cross-modal comparison. Radar off-centre Doppler energy vs.\ independently measured IMU vibration energy, one point per frame, pooled over the healthy and two vibration captures (Spearman $\rho=0.41$). Frames are temporally correlated within only three condition-level captures; the plot therefore supports a between-condition trend, not a capture-level significance claim.}
  \label{fig:crossmodal}
\end{figure}

\section{Per-Fault Error Analysis under Cross-Severity Shift}
\label{app:confusion}

\begin{figure}[tbp]
  \centering
  \includegraphics[width=\linewidth]{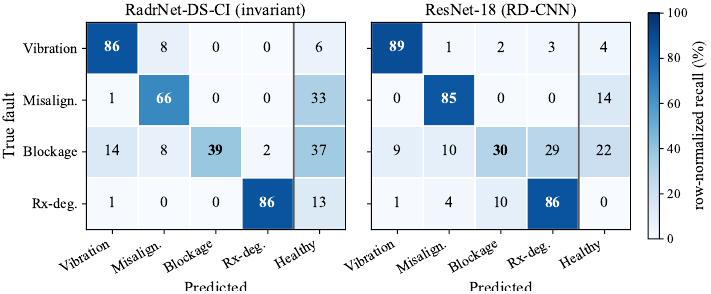}
  \caption{\textbf{Cross-severity per-fault confusion.} Row-normalized recall ($\%$) for RadrNet-DS-CI (left) and ResNet-18 (right), pooled over three seeds and both fold directions at the full chirp budget ($k{=}64$). The true classes are the four induced faults (healthy has no severity to cross); the separated \emph{Healthy} column counts frames the five-way model misreads as healthy.}
  \label{fig:crosssev_confusion}
\end{figure}

To show \emph{where} the cross-severity gap comes from, Figure~\ref{fig:crosssev_confusion} reports row-normalized confusion matrices for the capture-invariant headline model (RadrNet-DS-CI) and the range--Doppler CNN baseline (ResNet-18), pooled over all three seeds and both fold directions at the full chirp budget ($k{=}64$, $n{=}4800$ test predictions per model). The four induced faults are the true classes (healthy has no severity to cross), but the five-way classifiers can still predict \emph{healthy}, so that column is shown separately. Vibration is recovered well by both models ($86$--$89\%$). The dominant failure mode is collapse to \emph{healthy}: a third of misalignment frames read as healthy for the invariant model ($66\%$ recall, vs.\ $85\%$ for the RD-CNN), and over a third of blockage frames for both, consistent with these faults being subtle in the averaged power map. Blockage is the hardest class ($39\%$ recall for the invariant model, $30\%$ for the RD-CNN), and the RD-CNN additionally confuses it with receive degradation ($29\%$); both recover receive degradation well ($86\%$). The blockage- and misalignment-to-healthy confusion is the principal headroom in the benchmark. These matrices are computed from the final-epoch weights of the prefix-evaluation path (Section~\ref{app:ci_anytime}), whereas the per-fault recalls in Table~\ref{tab:ci_ablation} come from the best-epoch records behind Table~\ref{tab:crosssev_full}; the latter are therefore higher (e.g.\ $56.8\%$ vs.\ $39\%$ blockage recall for RadrNet-DS-CI), and the two views should not be mixed.

\section{Additional Analyses}
\label{app:extra}

\begin{figure}[tbp]
  \centering
  \includegraphics[width=0.82\linewidth]{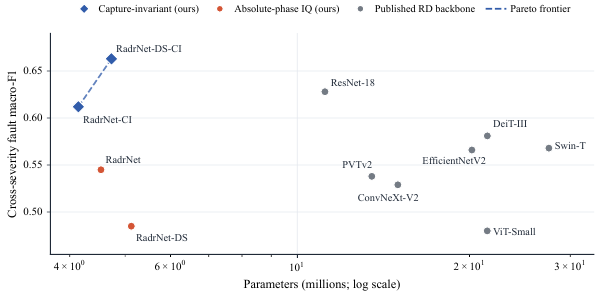}
  \caption{\textbf{Accuracy--efficiency trade-off.} Controlled cross-severity fault macro-F1 (three-seed mean) vs.\ model size (parameters, log scale). Marker colour denotes model family (capture-invariant, absolute-phase raw-IQ, RD-map backbone); the dashed line is the Pareto frontier.}
  \label{fig:pareto}
\end{figure}

\noindent\textbf{Accuracy--efficiency trade-off.} Figure~\ref{fig:pareto} plots controlled cross-severity fault macro-F1 against parameter count. The capture-invariant models are Pareto-optimal: RadrNet-DS-CI attains the top score ($0.663$) at $4.7$\,M parameters, under half the size of ResNet-18 ($11$\,M) and roughly a fifth of the transformer backbones ($20$--$28$\,M), all of which it outperforms. RadrNet-CI is the smallest model overall ($4.1$\,M) yet still beats every RD-map backbone except ResNet-18. Parameter count alone therefore does not explain the cross-severity ranking.

\begin{figure*}[tp]
  \centering
  \includegraphics[width=\linewidth]{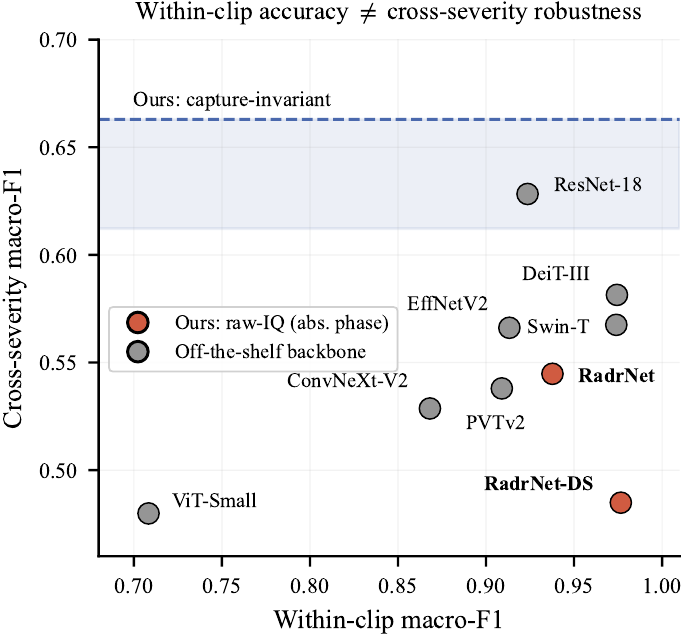}
  \caption{\textbf{Within-clip vs.\ cross-severity macro-F1 per model.} Each point is a model at its within-clip fault macro-F1 ($x$, from \texttt{v1\_within\_clip\_mimo*.json}) and three-seed cross-severity F1 ($y$, the main cross-severity table values). Marker colour denotes raw-IQ (orange) vs.\ RD-map (grey); the shaded band marks the capture-invariant models.}
  \label{fig:collapse}
\end{figure*}

\noindent\textbf{Within-clip vs.\ cross-severity: a re-ranking.} Figure~\ref{fig:collapse} contrasts each model's within-clip and controlled cross-severity fault macro-F1. Within clip, most models near-saturate ($0.87$--$0.98$; the weakest, ViT-Small, is at $0.71$), with the raw-IQ models among the leaders. Under cross-severity shift, every model declines and the absolute-phase raw-IQ models fall furthest (RadrNet-DS $0.977\!\to\!0.485$), dropping below the RD-map CNNs they previously led. This re-ranking is consistent with capture-specific phase and gain acting as within-clip shortcuts. The capture-invariant models recover the top of the cross-severity ranking.

\section{Compute Resources}
\label{app:compute}

The benchmark was run end-to-end on a single Windows-11 workstation with two NVIDIA RTX A4000 GPUs (16\,GB each), accessed via WSL2 / Ubuntu 22.04. Only one GPU is used at a time. Approximate compute budget:

\begin{itemize}[leftmargin=*,itemsep=2pt,topsep=2pt]
\item One training cache build (raw .bin $\to$ HDF5 with RD maps + IQ): $\sim$11\,min for single-Tx, $\sim$13\,min for full 192-channel MIMO cache.
\item Per-model within-clip MIMO training: $\sim$5\,min for ResNet-18, $\sim$7\,min for the transformer backbones (ViT-Small, DeiT-III, Swin-T, PVTv2), $\sim$5--7\,min for ConvNeXt-V2 / EfficientNetV2, and $\sim$25\,min for the RadrNet family (the dominant cost is the Mamba selective scan with checkpointing).
\item Anytime benchmark: five seeds $\times$ 2 models (ResNet-18, RadrNet-DS) at 20 epochs/seed, $\sim$5\,GPU-hours; the chirp-budget evaluation sweep itself is inference-only.
\item Few-shot benchmark: 15-epoch base training plus 30-step adaptation per support draw across the label budgets, $<$2\,GPU-hours.
\item Controlled cross-severity benchmark (2 folds per run; 25 epochs for ConvNeXt-V2, DeiT-III, Swin-T, PVTv2, EfficientNetV2 and the single-stream RadrNet, 20 for ResNet-18, ViT-Small, RadrNet-DS and the capture-invariant runs): three seeds for all 11 models in the reported ranking, plus the prefix-budget evaluations and CI few-shot run, for $\sim$10\,GPU-hours.
\item Total project compute including iteration, ablations, the published baselines, and the capture-invariant cross-severity runs: $<$25\,GPU-hours.
\end{itemize}

One command reproduces the within-clip benchmark: \nolinkurl{python scripts/train_within_clip_mimo.py --cache .../training_cache.h5}. The scripts \nolinkurl{aggregate_anytime.py} and \nolinkurl{aggregate_fewshot.py} aggregate the two budget studies. Individual jobs complete within an hour on an A4000; the full multi-seed anytime sweep takes approximately 5 GPU-hours.

\section{Capture Protocol Details (Lab Playbook Excerpts)}
\label{app:protocol}

The code repository provides the full capture protocol, bill of materials, severity-gate measurements, per-clip metadata schema, and printable checklist (\nolinkurl{docs/FAULT_INDUCTION_SOP.md}; \nolinkurl{docs/FAULT_INDUCTION_CHECKLIST.pdf}). Table~\ref{tab:induction} summarises the benchmark settings.

\begin{table*}[tp]
\centering
\caption{Per-fault induction details for the captures.}
\label{tab:induction}
\small
\setlength{\tabcolsep}{3pt}
\begin{tabular}{p{1.6cm}p{4.0cm}p{2.3cm}p{2.3cm}}
\toprule
\textbf{Fault} & \textbf{Hardware} & \textbf{S1 (mild)} & \textbf{S2 (severe)} \\
\midrule
Vibration       & Eccentric-mass DC motor, bolted to radar bracket, BNO055 IMU records ground-truth tri-axial acceleration & 20\,Hz drive frequency & 40\,Hz drive frequency \\
\midrule
Misalign.\ & Custom precision yaw-offset jig, 1$^\circ$ graduations, locked with set-screw, verified with 0.1$^\circ$ digital inclinometer & 5$^\circ$ yaw  & 10$^\circ$ yaw \\
\midrule
Blockage        & $1/16$-inch clear polycarbonate sheet on top of the always-installed $1/16$-inch polypropylene radome; laser-cut coverage templates, glossy face inward & 30\% coverage of the radar aperture & 60\% coverage \\
\midrule
Rx degradation  & 3M~1181 copper foil tape (0.5-inch wide, 25\,$\mu$m thick) fully covering targeted Rx antenna patches; per-channel dB drop verified by single-tone calibration sweep before each capture & 6 of 16 Rx channels covered & 10 of 16 Rx channels covered \\
\bottomrule
\end{tabular}
\end{table*}

\noindent\textbf{Capture session protocol.}
Each session: (1) power on and thermally stabilise for 10 min; (2) record a healthy baseline; (3) induce faults in a session-seeded random order; (4) for each fault, measure the severity gate, record 5 min, remove the fixture, and record a 5 s recovery reference; (5) record a closing healthy reference; and (6) run per-file SHA-256 and automated QA via \nolinkurl{src/io/qa.py}. Randomisation reduces temporal-position confounding.

\section{Future Directions}
\label{app:roadmap}

Future releases aim to broaden Rad-R along several axes: additional severity levels, multi-day / multi-device / multi-scene capture, further fault and interference modalities, and additional model baselines. Concrete specifications will accompany those releases.

\section{Broader Impact}
\label{app:broader_impact}

\noindent\textbf{Positive.} Better automotive-radar fault diagnosis could support earlier detection of sensor degradation in ADAS and autonomous-driving systems. Releasing raw ADC under an open license lowers the barrier to academic radar-perception research, historically constrained by proprietary data and expensive hardware. The fault-induction SOP and printable checklist standardise the capture protocol so future researchers can extend Rad-R reproducibly.

\noindent\textbf{Negative / risks.}
\emph{(a) Camera privacy:} the two co-recorded camera streams during outdoor captures may incidentally capture pedestrians or license plates. Every released camera frame is processed through an automated PII-redaction pipeline that Gaussian-blurs every detected human face and licence plate before upload; the unredacted originals never leave the authors' workstation, and a 5\%-per-capture random sample is manually verified for missed detections.
\emph{(b) Adversarial misuse:} a fault classifier could in principle be inverted into a fault-induction recipe. We judge this risk low: the released fault-induction procedures are already public, the underlying physics is well known to anyone with a 77\,GHz radar, and adversarial induction requires physical access to the vehicle. We do not release production-deployment detector weights; the released weights are research checkpoints only.
\emph{(c) Over-claiming generalisation:} practitioners might read these numbers as production-ready. We mitigate this by treating within-clip F1 as an upper bound, framing the dataset as a single-day / single-device research dataset, and committing to broaden coverage in future releases (Appendix~\ref{app:roadmap}).

\section{Reproducibility and Responsible-AI Summary}
\label{app:repro}

This section consolidates the reproducibility and responsible-AI details supporting the main-text claims.

\noindent\textbf{Reproducibility.} All models share the same core optimizer, learning rate, weight decay, batch size, cosine schedule, and gradient clipping; protocol-specific epoch budgets and seed counts are stated in Appendix~\ref{app:hyperparams}. The appendix also records the RadrNet configuration and capture-invariance augmentations. The within-clip split definition, cache-build script, training/evaluation scripts, and result JSONs are included in the release so that each reported protocol can be reconstructed from its stated configuration.

\noindent\textbf{Data and code access.} The code release includes training and evaluation code, result JSONs, Croissant 1.0 metadata, and a compact PII-redacted data sample. The complete dataset and code will be released publicly; data will be licensed under CC~BY~4.0 and code under the licenses stated above. The metadata fields cover data collection, annotation protocol, maintenance, personal/sensitive information, limitations, biases, use cases, social impact, sources, provenance, and preprocessing.

\noindent\textbf{Statistical significance.} The anytime benchmark reports five-seed mean$\pm$std with BCa-95\% bootstrap CIs, one-sided paired-bootstrap $p$-values ($n{=}2000$) and McNemar exact $p$-values on 1\,800 pooled samples; the few-shot benchmark uses a one-sided unpaired bootstrap ($n{=}2000$) over the support draws, with RadrNet-DS and ResNet-18 carried across three seeds and DeiT-III reported at a single seed.

\noindent\textbf{Compute.} The full benchmark runs on two NVIDIA RTX A4000 GPUs (16\,GB; one used at a time) under WSL2 / Ubuntu 22.04, CUDA 11.8, PyTorch 2.5.1, \texttt{mamba\_ssm} 2.3.1. Per-model within-clip wall-clock: $\sim$5\,min ResNet-18, $\sim$7\,min the transformer backbones, $\sim$25\,min RadrNet; the anytime benchmark adds $\sim$5\,GPU-hours (five seeds, 2 models) and the few-shot benchmark $<$2\,GPU-hours; total project compute $<$25 GPU-hours; cache build $\sim$11--13\,min.

\noindent\textbf{Research ethics.} No human-subject, medical, or deliberately collected sensitive personal data. Incidental pedestrians / plates in outdoor camera frames are redacted as above. All hardware was operated on private lots and permitted campus roads at low speed; no public-road testing of fault-induced sensors occurred. Fault-induction fixtures use commercial off-the-shelf hardware and are fully reversible.

\end{document}